%% file: main.tex
\documentclass[11pt]{article}

\usepackage[preprint]{acl}
\usepackage{hwemoji}

\usepackage{times}
\usepackage{amsmath,amssymb}
\usepackage{latexsym}
\usepackage{booktabs}
\usepackage{multirow}
\usepackage[T1]{fontenc}
\usepackage[utf8]{inputenc}
\usepackage{microtype}
\usepackage{inconsolata}
\usepackage{graphicx}
\usepackage{dblfloatfix} 
\usepackage{float}        
\usepackage{caption}
\usepackage{tablefootnote}
\usepackage{graphicx}
\usepackage{subcaption}

\usepackage{pifont}

\newcommand{\na}{n/a}

\title{BEAR-Bench: A Bilingual Enterprise and Academic
       Reasoning Benchmark for Multimodal Models}

\author{
\textbf{Liubov Chubarova\textsuperscript{1}},
 \textbf{Alexandra Kuleshova\textsuperscript{2}},
 \textbf{Daniil Volkov$^*$\textsuperscript{2}},
 \textbf{Kirill Sultanov$^*$\textsuperscript{1}},
 \textbf{Alexey Zaytsev\textsuperscript{2}}
\\
\\
  \textsuperscript{1} Yandex \textsuperscript{2} Applied AI Institute
\\
  \small{    \textbf{Correspondence:} \href{mailto:email@domain}{bazarovaai.239@gmail.com}
 }
}

\begin{document}

\maketitle

\def\thefootnote{*}\footnotetext{Equal contribution.}

\begin{abstract}
While Multimodal Large Language Models (MLLMs) have made
significant strides in visual comprehension, their ability to
reason about text-dense, professional documents remains incompletely evaluated. Existing benchmarks emphasize information extraction, require external domain knowledge, or cover professional documents only as one of many settings. They are also largely English- or Chinese-centric, leaving other languages and Russian, in particular, substantially underrepresented. To address these limitations, we
introduce \textbf{BEAR-Bench} (\textbf{B}ilingual
\textbf{E}nterprise and \textbf{A}cademic \textbf{R}easoning),
a self-contained, complex English-and-Russian benchmark comprising
1000 human-annotated questions based
on text-rich business and scientific documents.
We evaluate 16 proprietary and
open-weight MLLMs, including Gemini 3.1 Pro and Qwen3.5-397B,
on BEAR-Bench and observe clear headroom even for the strongest
systems. Finally, we use the resulting model outputs to compare
existing hallucination detection methods, evaluating not only
how often models fail on BEAR-Bench but also how reliably
those failures can be identified.
\end{abstract}

\input{intro}

\input{related_work}
\input{benchmark}
\input{statistics}
\input{experiments}
\input{conclusion}

\bibliography{custom}

\appendix

\section{Error taxonomy}\label{app:error_types}

Below, we elaborate on the error taxonomy derived in our study.
\begin{enumerate}
    \item \textbf{C1 — Spatial localization and object matching.}  
    The model incorrectly identifies where objects are located in the image and how they relate to each other (e.g., selecting a neighboring element, misreading whether a marker lies inside a region, boundary intersections, label-to-object matching, arrow direction, or links between blocks).

    \item \textbf{C2 — Counting and aggregation errors of visual elements.}  
    The model makes mistakes when counting objects or aggregating extracted elements (e.g., points, circles, arrows, rows, columns, people, links, paths, labels, or table values), including missed elements, extra elements, or   incorrect summation.

    \item \textbf{C3 — Errors in reading text, numbers, and visual attributes.}  
    The model incorrectly reads text, numbers, symbols, or small labels (OCR-related issues), and may also misidentify visual attributes such as color, marker shape, text style, italics/boldface, or legend encodings.

    \item \textbf{C4 — Errors in extracting values from charts.}  
    The model incorrectly reads quantitative values from plots/graphs (e.g., axis scale, ticks, point coordinates, values at specific \(x\), peaks, minima, maxima, plateaus, trends, or ranges).

    \item \textbf{C5 — Semantic, instruction-following, and logical/arithmetic errors.}  
    The model misunderstands task conditions, categories, terms, units, or filtering rules, and/or makes reasoning or arithmetic mistakes after extraction (e.g., wrong interpretation, entity confusion, incorrect formulas, hallucinated assumptions, or incomplete answers).
\end{enumerate}

Representative examples of each error type are visualized in Figures~\ref{fig:err-c1}-\ref{fig:err-c5}.

\section{Illustrative items from MWS Vision Bench}
\label{app:mws-examples}

\begin{figure*}[t]
  \centering
  \includegraphics[width=\textwidth]{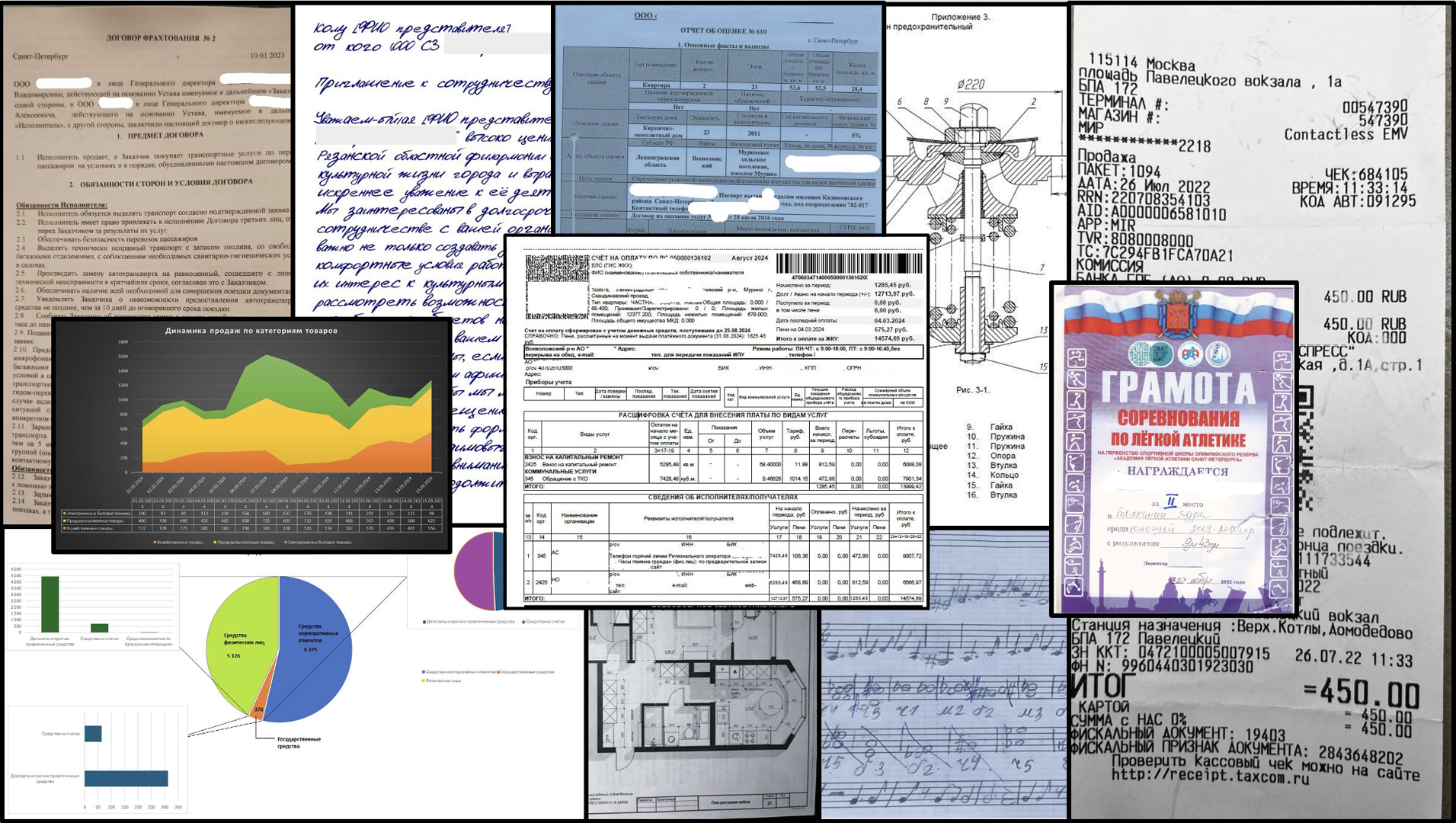}
  \caption{Illustrative items from the public validation split of MWS Vision Bench~\cite{mwsvisionbench2025}. The mix of personal handwriting, receipts, and document-processing tasks differs from BEAR-Bench's professional, text-dense pages.}
  \label{fig:mws-examples}
\end{figure*}

Figure~\ref{fig:mws-examples} shows public validation items from MWS Vision Bench~\cite{mwsvisionbench2025}. The illustration is taken from the dataset's Hugging Face page.\footnote{\url{https://huggingface.co/datasets/MTSAIR/MWS-Vision-Bench}} We include them to make the contrast with BEAR-Bench concrete: MWS mixes business scans with personal handwriting, receipts, and form-style pages, and a large share of its tasks are OCR, grounding, and key-information extraction. BEAR-Bench instead targets multi-step questions on text-dense scientific and business document pages (Figure~\ref{fig:examples}). 

\section{Accuracy vs Reasoning Depth}
\begin{figure}[t]
    \centering
    \includegraphics[width=\linewidth,height=0.22\textheight,keepaspectratio]{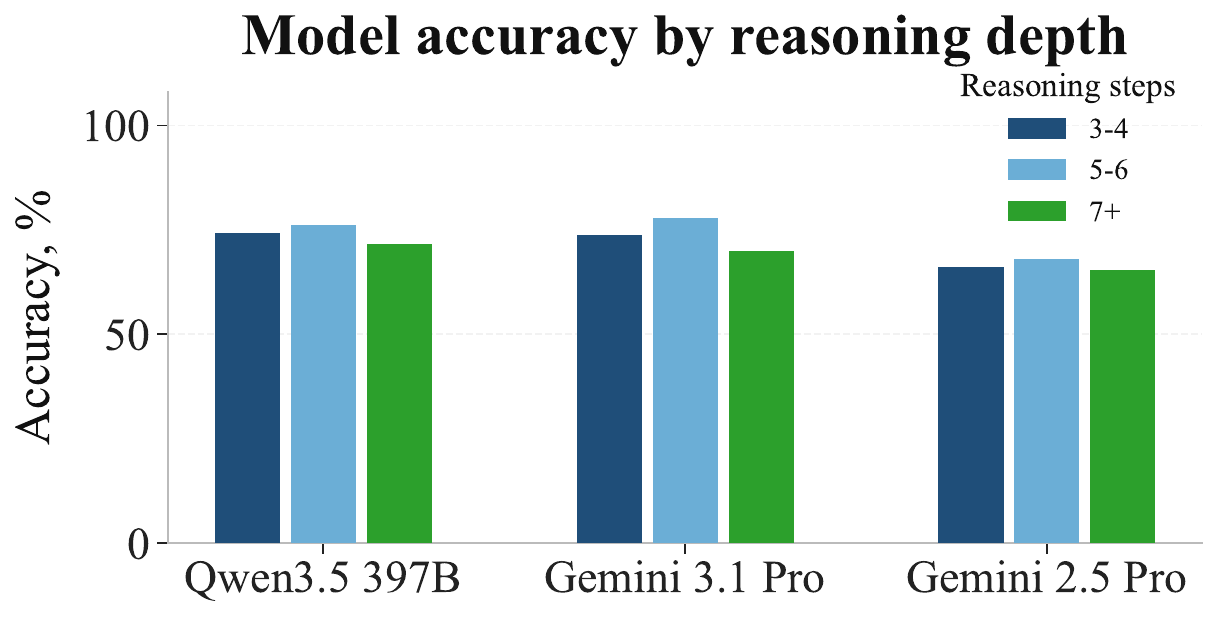}
    \caption{Accuracy on BEAR-Bench as a function of reasoning depth (number of steps per item) for three frontier models.}
    \label{fig:accuracy_vs_reasoning_count}
\end{figure}

Figure~\ref{fig:accuracy_vs_reasoning_count} demonstrates the accuracy broken down by annotated reasoning-step count for several proprietary models. The trend is non-monotonic, suggesting that on knowledge-free tasks, frontier models are constrained more by visual perception than by the ability to execute long reasoning chains.

\begin{figure}[t]
  \centering
  \includegraphics[width=\linewidth,height=0.22\textheight,keepaspectratio]{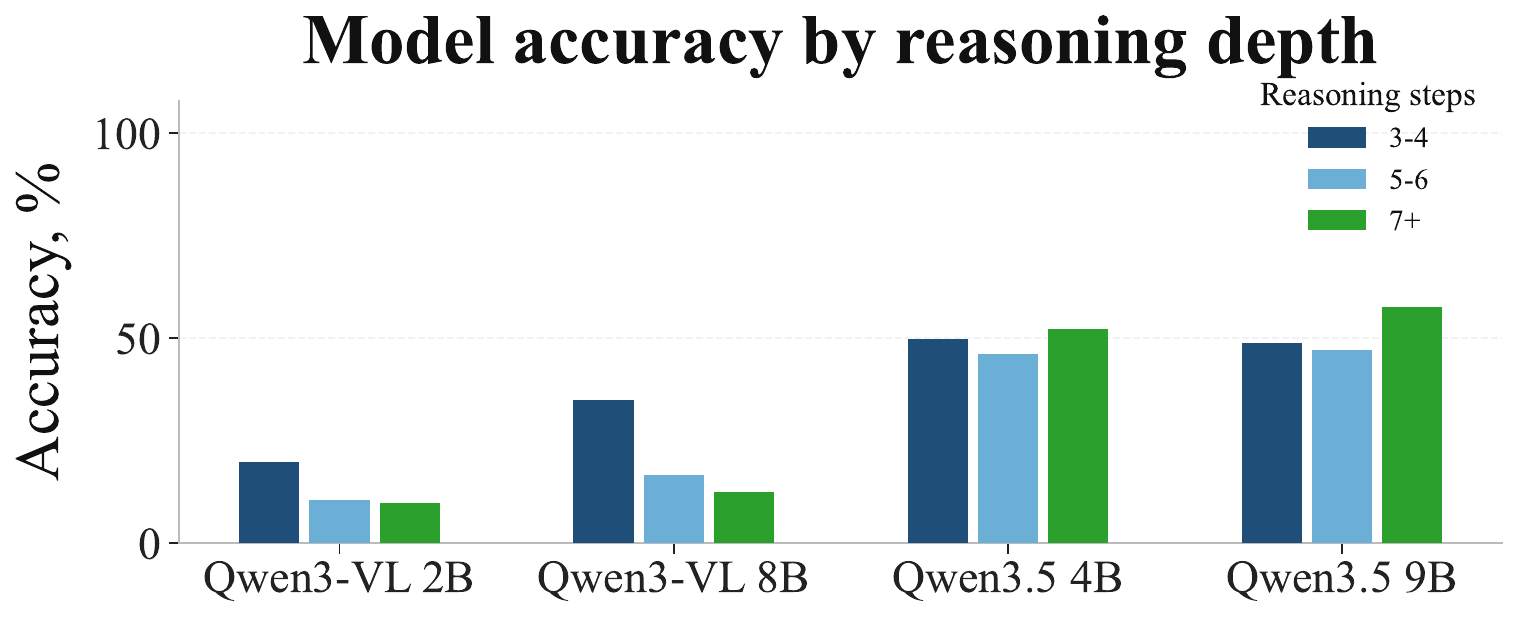}
  \caption{Accuracy on BEAR-Bench as a function of reasoning depth (number of steps per item) for four open-weight models.}
  \label{fig:accuracy_vs_reasoning_count_opensource}
\end{figure}

We repeat this analysis for four open-weight models (Figure~\ref{fig:accuracy_vs_reasoning_count_opensource}), where the two model families exhibit contrasting behaviour. The accuracy of the reasoning-tuned Qwen3.5 models remains stable as the annotated reasoning depth grows, whereas the instruction-tuned Qwen3-VL models degrade on items requiring seven or more steps. A likely reason is that reasoning-tuned models are trained to produce long chains of reasoning, so questions that need more steps cost them little extra accuracy, while instruction-tuned models are not trained for this and lose accuracy as more steps are needed. 

\section{Additional data statistics: response length}
\label{app:response_length}
Table~\ref{tab:median_response_length_words} reports the median response length in words for each evaluated model. Lengths are computed by whitespace-splitting each model's stored answer field. Proprietary and open-weight models differ sharply: several API systems return very short answers (median 2--3 words), while reasoning-oriented open-weight models often produce long intermediate chains (median above 1{,}000 words).

\begin{table}[htbp]
  \centering
  \scriptsize
  \caption{Median response length (words) by model. Words are whitespace-split tokens from each model's stored \texttt{answer} field.}
  \label{tab:median_response_length_words}
  \begin{tabular}{llr}
    \toprule
    Group & Model & Median words \\
    \midrule
    \multirow{8}{*}{Proprietary / API}
      & Gemini~3.1 Pro & 3 \\
      & Gemini~3.1 Flash & 46 \\
      & Gemini~2.5 Pro & 3 \\
      & Gemini~2.5 Flash & 123 \\
      & Qwen~3.6 Plus & 2 \\
      & Qwen3.5 397B A17B & 2 \\
      & Claude Sonnet~4.6 & 136 \\
      & Claude Opus~4.6 & 149 \\
    \midrule
    \multirow{8}{*}{Open-weight}
      & Qwen3.5-0.8B & 96 \\
      & Qwen3.5-27B & 1036 \\
      & Qwen3.5-4B & 1206 \\
      & Qwen3.5-9B & 1162 \\
      & Qwen3-VL-2B-Instruct & 117 \\
      & Qwen3-VL-4B-Thinking & 1085 \\
      & Qwen3-VL-8B-Instruct & 2 \\
      & gemma-4-31B-it & 12 \\
    \bottomrule
  \end{tabular}
\end{table}

\section{Hallucination Detection Metrics}
\label{sec:appendix}

This appendix reports the full 5-fold cross-validation results
for the hallucination detectors evaluated in
Section~\ref{sec:hallucination}. We compare uncertainty-based
scores, ContextualLens, SUQ Probe, VIB-Probe, FaithScan, and an MLLM-as-a-judge
baseline under two regimes: proxy-based detection for
proprietary models and native-signal detection for open-weight
models. Tables~\ref{tab:5fold_gemini} and~\ref{tab:5fold_qwen_claude}
present AUROC, AUC-PR, and balanced accuracy for Gemini, Qwen,
and Claude outputs; Table~\ref{tab:5fold_openweight} reports
the corresponding results for Qwen3.5-9B and Qwen3.5-27B.

\begin{table}[htbp]
\centering
\small
\caption{Hallucination detection (5-fold CV): Gemini models. AUROC and AUC-PR are omitted for VLM-as-judge because the judge returns binary verdicts.}
\label{tab:5fold_gemini}
\resizebox{\linewidth}{!}{%
\begin{tabular}{lccc}
\toprule
\textbf{Method} & \textbf{AUROC} & \textbf{AUC-PR} & \textbf{BalAcc} \\
\midrule
\multicolumn{4}{c}{\textbf{Gemini 3.1 Pro}} \\
\midrule
Max prob & 0.64 & 0.38 & 0.60 \\
Avg prob & 0.48 & 0.26 & 0.48 \\
Log-likelihood & 0.50 & 0.29 & 0.46 \\
Max entropy & 0.60 & 0.41 & 0.52 \\
Avg entropy & 0.50 & 0.26 & 0.52 \\
Perplexity & 0.50 & 0.29 & 0.46 \\
ContextualLens & 0.41 & 0.24 & 0.40 \\
SUQ Probe & \underline{0.76} & 0.54 & \underline{0.69} \\
VIB-Probe & \textbf{0.79} & \textbf{0.62} & \textbf{0.71} \\
FaithScan & \underline{0.76} & \underline{0.56} & 0.68 \\
VLM-as-judge & {--} & {--} & 0.57 \\
Random & 0.50 & 0.27 & 0.50 \\
\midrule
\multicolumn{4}{c}{\textbf{Gemini 3.1 Flash}} \\
\midrule
Max prob & 0.50 & 0.45 & 0.49 \\
Avg prob & 0.69 & 0.59 & 0.65 \\
Log-likelihood & 0.68 & 0.56 & 0.62 \\
Max entropy & 0.61 & 0.52 & 0.59 \\
Avg entropy & 0.71 & 0.60 & 0.66 \\
Perplexity & 0.68 & 0.56 & 0.62 \\
ContextualLens & 0.56 & 0.47 & 0.57 \\
SUQ Probe & 0.75 & 0.68 & \underline{0.70} \\
VIB-Probe & \underline{0.77} & \underline{0.71} & \underline{0.70} \\
FaithScan & \textbf{0.79} & \textbf{0.74} & \textbf{0.72} \\
VLM-as-judge & {--} & {--} & 0.69 \\
Random & 0.50 & 0.42 & 0.50 \\
\midrule
\multicolumn{4}{c}{\textbf{Gemini 2.5 Pro}} \\
\midrule
Max prob & 0.63 & 0.45 & 0.61 \\
Avg prob & 0.54 & 0.38 & 0.51 \\
Log-likelihood & 0.57 & 0.40 & 0.53 \\
Max entropy & 0.58 & 0.47 & 0.52 \\
Avg entropy & 0.54 & 0.37 & 0.54 \\
Perplexity & 0.57 & 0.40 & 0.53 \\
ContextualLens & 0.46 & 0.35 & 0.46 \\
SUQ Probe & 0.77 & 0.65 & 0.70 \\
VIB-Probe & \textbf{0.82} & \textbf{0.73} & \textbf{0.74} \\
FaithScan & \underline{0.80} & \underline{0.70} & \underline{0.71} \\
VLM-as-judge & {--} & {--} & 0.60 \\
Random & 0.50 & 0.34 & 0.50 \\
\midrule
\multicolumn{4}{c}{\textbf{Gemini 2.5 Flash}} \\
\midrule
Max prob & 0.50 & 0.53 & 0.49 \\
Avg prob & 0.71 & 0.72 & 0.65 \\
Log-likelihood & 0.71 & 0.71 & 0.65 \\
Max entropy & 0.54 & 0.57 & 0.52 \\
Avg entropy & 0.70 & 0.71 & 0.65 \\
Perplexity & 0.71 & 0.71 & 0.65 \\
ContextualLens & 0.60 & 0.60 & 0.58 \\
SUQ Probe & \underline{0.80} & \underline{0.81} & \underline{0.74} \\
VIB-Probe & \textbf{0.82} & \textbf{0.82} & \textbf{0.75} \\
FaithScan & 0.79 & \underline{0.81} & 0.72 \\
VLM-as-judge & {--} & {--} & 0.72 \\
Random & 0.50 & 0.52 & 0.50 \\
\bottomrule
\end{tabular}}
\end{table}

\begin{table}[htbp]
\centering
\small
\caption{Hallucination detection (5-fold CV): Qwen and Claude models. AUROC and AUC-PR are omitted for VLM-as-judge because the judge returns binary verdicts.}
\label{tab:5fold_qwen_claude}
\resizebox{\linewidth}{!}{%
\begin{tabular}{lccc}
\toprule
\textbf{Method} & \textbf{AUROC} & \textbf{AUC-PR} & \textbf{BalAcc} \\
\midrule
\multicolumn{4}{c}{\textbf{Qwen 3.6 Plus}} \\
\midrule
Max prob & 0.72 & 0.61 & 0.66 \\
Avg prob & 0.62 & 0.48 & 0.56 \\
Log-likelihood & 0.67 & 0.58 & 0.57 \\
Max entropy & 0.62 & 0.50 & 0.56 \\
Avg entropy & 0.59 & 0.44 & 0.57 \\
Perplexity & 0.67 & 0.58 & 0.57 \\
ContextualLens & 0.48 & 0.39 & 0.47 \\
SUQ Probe & 0.80 & 0.72 & 0.72 \\
VIB-Probe & \textbf{0.84} & \textbf{0.78} & \textbf{0.76} \\
FaithScan & \underline{0.82} & \underline{0.76} & \underline{0.73} \\
VLM-as-judge & {--} & {--} & 0.62 \\
Random & 0.50 & 0.37 & 0.50 \\
\midrule
\multicolumn{4}{c}{\textbf{Qwen3.5 397B}} \\
\midrule
Max prob & 0.62 & 0.38 & 0.59 \\
Avg prob & 0.49 & 0.27 & 0.48 \\
Log-likelihood & 0.51 & 0.28 & 0.48 \\
Max entropy & 0.57 & 0.38 & 0.51 \\
Avg entropy & 0.50 & 0.27 & 0.52 \\
Perplexity & 0.51 & 0.28 & 0.48 \\
ContextualLens & 0.45 & 0.26 & 0.46 \\
SUQ Probe & \underline{0.76} & \underline{0.57} & \underline{0.67} \\
VIB-Probe & \textbf{0.80} & \textbf{0.65} & \textbf{0.72} \\
FaithScan & 0.75 & 0.56 & \underline{0.67} \\
VLM-as-judge & {--} & {--} & 0.57 \\
Random & 0.50 & 0.27 & 0.50 \\
\midrule
\multicolumn{4}{c}{\textbf{Claude Sonnet 4.6}} \\
\midrule
Max prob & 0.50 & 0.39 & 0.50 \\
Avg prob & 0.73 & 0.61 & 0.65 \\
Log-likelihood & 0.70 & 0.56 & 0.66 \\
Max entropy & 0.73 & 0.65 & 0.67 \\
Avg entropy & 0.75 & 0.65 & 0.69 \\
Perplexity & 0.70 & 0.56 & 0.66 \\
ContextualLens & 0.51 & 0.39 & 0.53 \\
SUQ Probe & \underline{0.81} & \underline{0.75} & \underline{0.74} \\
VIB-Probe & \textbf{0.84} & \textbf{0.78} & \textbf{0.75} \\
FaithScan & \underline{0.81} & \underline{0.75} & 0.72 \\
VLM-as-judge & {--} & {--} & 0.68 \\
Random & 0.50 & 0.38 & 0.50 \\
\midrule
\multicolumn{4}{c}{\textbf{Claude Opus 4.6}} \\
\midrule
Max prob & 0.51 & 0.41 & 0.51 \\
Avg prob & 0.71 & 0.61 & 0.66 \\
Log-likelihood & 0.68 & 0.55 & 0.63 \\
Max entropy & 0.68 & 0.58 & 0.63 \\
Avg entropy & 0.73 & 0.63 & 0.68 \\
Perplexity & 0.68 & 0.55 & 0.63 \\
ContextualLens & 0.50 & 0.40 & 0.51 \\
SUQ Probe & 0.77 & 0.68 & 0.69 \\
VIB-Probe & \textbf{0.84} & \textbf{0.76} & \textbf{0.76} \\
FaithScan & \underline{0.81} & \underline{0.73} & \underline{0.73} \\
VLM-as-judge & {--} & {--} & 0.68 \\
Random & 0.50 & 0.40 & 0.50 \\
\bottomrule
\end{tabular}}
\end{table}

\begin{table}[htbp]
\centering
\small
\caption{Hallucination detection (5-fold CV): open-weight Qwen3.5 models (native signals). The VLM-as-a-judge results are computed over 997 valid samples for each model. AUROC and AUC-PR are omitted for VLM-as-judge because the judge returns binary verdicts.}
\label{tab:5fold_openweight}
\resizebox{\linewidth}{!}{%
\begin{tabular}{lccc}
\toprule
\textbf{Method} & \textbf{AUROC} & \textbf{AUC-PR} & \textbf{BalAcc} \\
\midrule
\multicolumn{4}{c}{\textbf{Qwen3.5-9B}} \\
\midrule
Max prob & 0.63 & 0.63 & 0.58 \\
Avg prob & 0.74 & 0.74 & 0.67 \\
Log-likelihood & 0.73 & 0.74 & 0.68 \\
Max entropy & 0.69 & 0.69 & 0.64 \\
Avg entropy & 0.74 & 0.75 & 0.68 \\
Perplexity & 0.73 & 0.74 & 0.68 \\
ContextualLens & 0.60 & 0.61 & 0.57 \\
SUQ Probe & 0.72 & 0.72 & 0.67 \\
VIB-Probe & \underline{0.76} & \underline{0.77} & 0.70 \\
FaithScan & \textbf{0.81} & \textbf{0.82} & \underline{0.73} \\
VLM-as-judge & {--} & {--} & \textbf{0.80} \\
Random & 0.50 & 0.51 & 0.50 \\
\midrule
\multicolumn{4}{c}{\textbf{Qwen3.5-27B}} \\
\midrule
Max prob & 0.65 & 0.56 & 0.60 \\
Avg prob & 0.70 & 0.61 & 0.63 \\
Log-likelihood & 0.70 & 0.61 & 0.63 \\
Max entropy & 0.69 & 0.61 & 0.63 \\
Avg entropy & 0.71 & 0.63 & 0.65 \\
Perplexity & 0.70 & 0.61 & 0.63 \\
ContextualLens & 0.63 & 0.56 & 0.59 \\
SUQ Probe & 0.65 & 0.55 & 0.60 \\
VIB-Probe & \underline{0.74} & \underline{0.65} & 0.68 \\
FaithScan & \textbf{0.82} & \textbf{0.77} & \underline{0.74} \\
VLM-as-judge & {--} & {--} & \textbf{0.81} \\
Random & 0.50 & 0.39 & 0.50 \\
\bottomrule
\end{tabular}}
\end{table}

\section{Chain-of-Thought Prompt}\label{app:cot-prompt}

\begin{figure}[t]
  \centering
  \includegraphics[width=\linewidth]{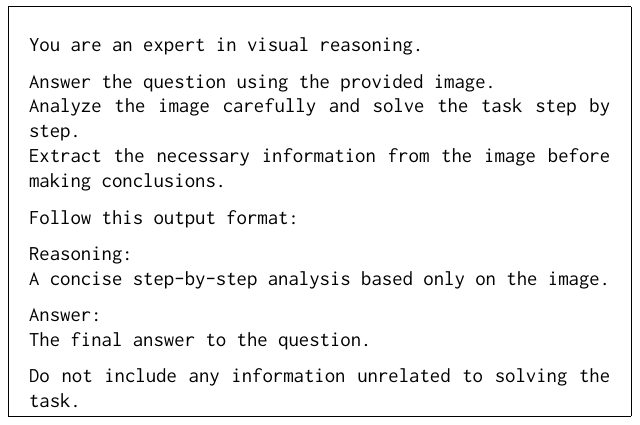}
  \caption{Chain-of-thought system prompt used in the CoT evaluation condition.}
  \label{fig:cot-prompt}
\end{figure}

Figure~\ref{fig:cot-prompt} shows the system prompt prepended to each image--question pair in the CoT condition reported in Table~\ref{tab:cot}.

\section{Human Annotation}

\subsection{Instructions for annotators}

\begin{figure*}[!t]
  \centering
  \includegraphics[width=\textwidth,height=0.78\textheight,keepaspectratio]{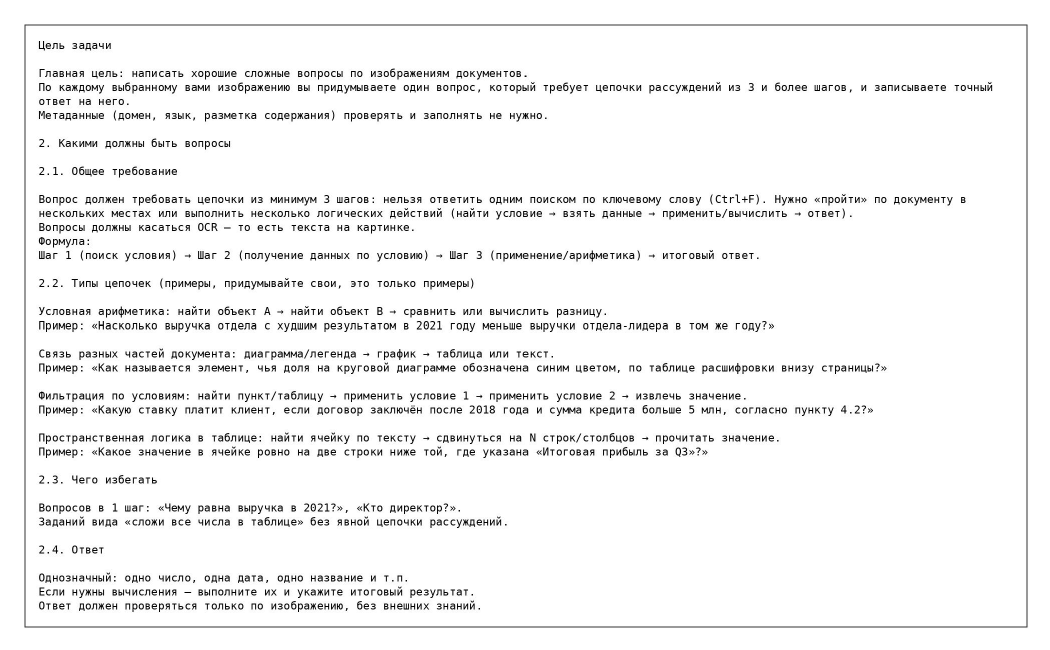}
  \caption{Original Russian instructions given to benchmark annotators. The excerpt states the annotation objective (one multi-step question per document image with a ground-truth answer) and Section~2, which defines question requirements: at least three reasoning steps, OCR-grounded text, exemplar chain types, items to avoid, and answer constraints. The translation to English is provided in Figure~\ref{fig:annotator-instructions-en}.}
  \label{fig:annotator-instructions-ru}
\end{figure*}

\begin{figure*}[!t]
  \centering
  \includegraphics[width=\textwidth,height=0.78\textheight,keepaspectratio]{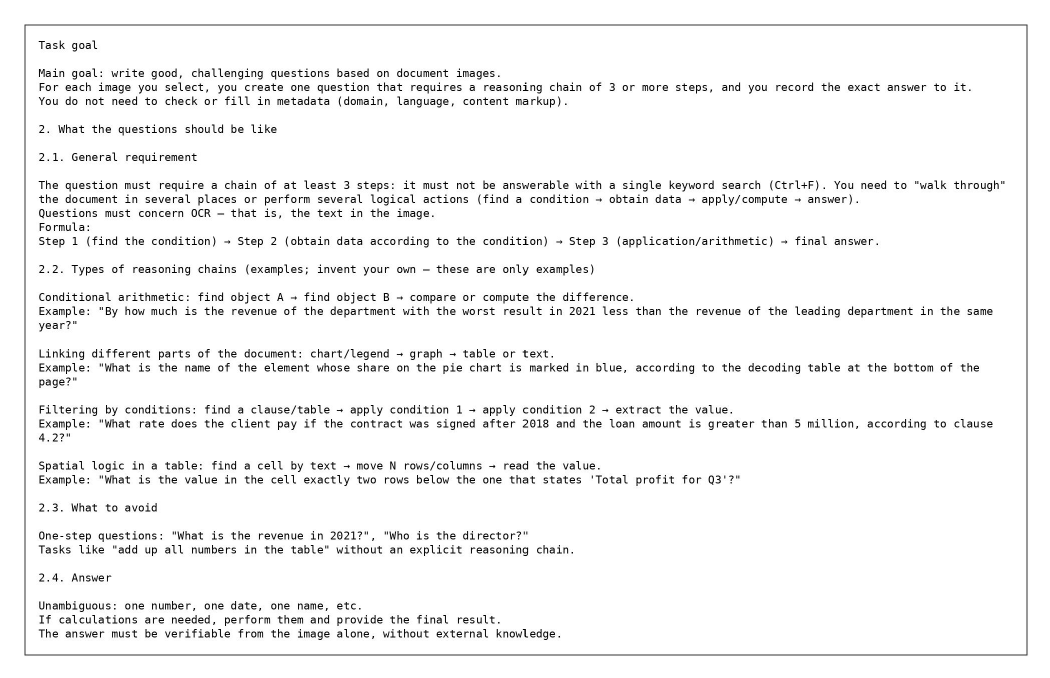}
  \caption{English translation of the annotator instructions shown in Figure~\ref{fig:annotator-instructions-ru}.}
  \label{fig:annotator-instructions-en}
\end{figure*}

The instructions given to annotators are shown in Figures~\ref{fig:annotator-instructions-ru} and \ref{fig:annotator-instructions-en} (the original text and its English translation, respectively). The main goal for the annotators was to design complex multi-hop questions that can be answered solely from the image content, without requiring any external expert knowledge. We introduced several examples of ``good'' and ``bad'' questions to elaborate on the task. The annotators were instructed to formulate the answers to the questions as briefly as possible (e.g., a single number).

\subsection{Recruitment \& payment}
Annotators were recruited through an open call posted in an internal student chat channel. Prior to the main task, each candidate completed a qualification test in which they generated 10 probe questions designed to elicit incorrect answers from Gemini 3.1 Pro. Candidates who successfully caused the model to fail on more than 4 out of 10 questions were selected to work on the dataset. Annotators were compensated at 4.4 times the Russian minimum wage.

\subsection{Ethics \& Consent}
No formal ethics review was required for this non-invasive annotation task. All participants provided informed consent.

\subsection{Demographics}
All annotators were aged 22–25 years and held at least a bachelor's degree in a technical field, with self-reported English proficiency at CEFR B2 or higher. The sample included 60\% male and 40\% female participants.

\section{Broader Impact, Data Use, and Compute Details}

\paragraph{Potential risks.}
BEAR-Bench is intended for research evaluation rather than autonomous decision making. Errors on its document-reasoning tasks can arise from misreading text, tables, figures, or equations and can consequently produce incorrect calculations or unsupported conclusions. If similar systems are used without human verification in financial, scientific, or other high-stakes workflows, such errors could lead to incorrect analyses or decisions. Performance also varies across languages and document types; therefore, aggregate benchmark scores should not be interpreted as evidence of reliable performance for every user population or document genre. We recommend using the benchmark for comparative evaluation and retaining human oversight in consequential settings.

\paragraph{Intended use.}
BEAR-Bench is designed primarily as an evaluation benchmark for multimodal reasoning and hallucination detection. It does not provide step-by-step rationale annotations, and therefore does not directly support training or evaluating explicit chain-of-thought reasoning. The benchmark should not be interpreted as a resource for certifying models for deployment in high-stakes financial, legal, or scientific decision-making settings.

\paragraph{Privacy and content.}
All source pages were obtained from publicly accessible official disclosure, government, intergovernmental, or academic sources, as described in Section~\ref{sec:benchmark}. We did not collect personal data directly from individuals and did not conduct additional content screening beyond selecting documents from these sources and verifying their licensing status. Public source documents may nevertheless contain names, affiliations, or other information present in the original records. We retain source attribution and recommend that users treat the benchmark as a research resource rather than as a source of information about individuals.

\paragraph{Compute infrastructure.}
Open-weight models were evaluated on an internal server with two NVIDIA H100 GPUs and five NVIDIA L40 GPUs. Proprietary models were accessed through the OpenRouter API; their underlying hardware configuration and parameter counts are not publicly available for all evaluated models. Reproducing the open-weight results requires approximately 100--120 GPU hours, whereas compute for proprietary models cannot be reliably estimated due to closed serving infrastructure.

\begin{figure*}[!t]
  \centering
  \includegraphics[width=\textwidth,height=0.78\textheight,keepaspectratio]{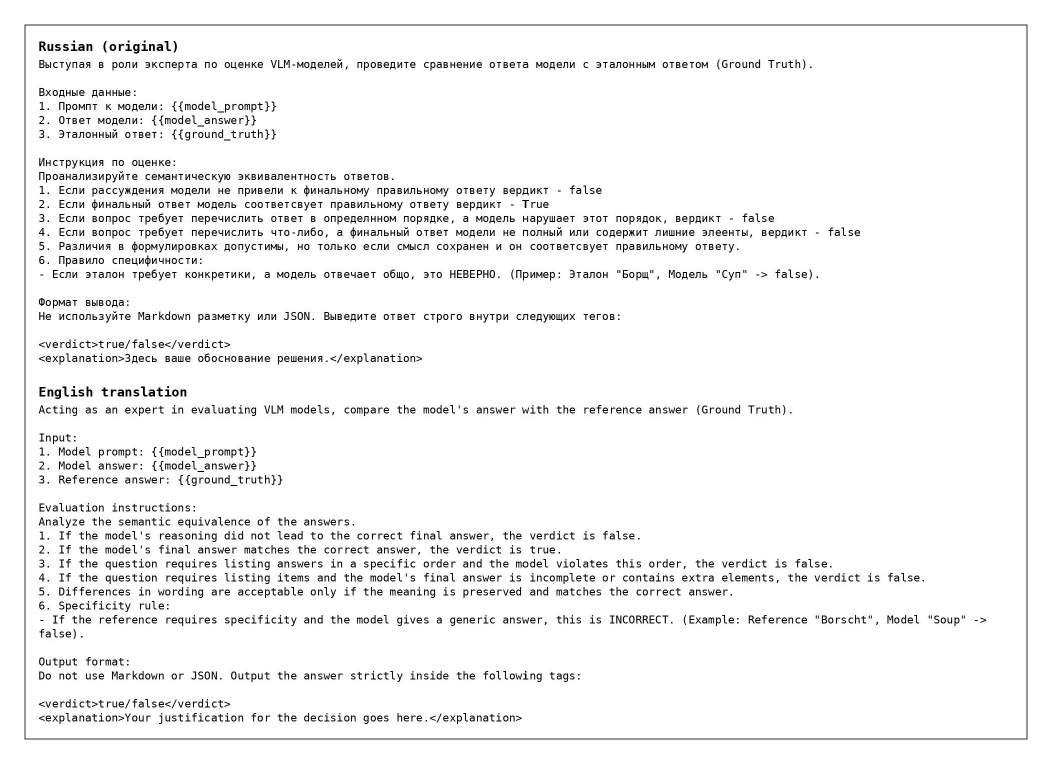}
  \caption{LLM-as-a-judge prompt used to score model responses against the ground truth.}
  \label{fig:llm-judge-prompt}
\end{figure*}

\section{Meta-feature predictors of correctness}
\label{app:metafeatures}

We study the impact of meta-features on the ability to predict whether a model will respond correctly. The study was run for each of the 16 models, covering a total of 9 features:
\texttt{domain\_science},
\texttt{lang\_ru},
\texttt{complexity},
\texttt{contains\_code},
\texttt{contains\_figures},
\texttt{contains\_diagrams},
\texttt{contains\_equations},
\texttt{question\_length},
and \texttt{answer\_length} (ground truth).

We run 4 complementary experiments: logistic regression with significance tests,
SHAP attribution, Pearson / point-biserial correlations, and testing the hypothesis that different features are considered significant for models of different sizes.
Before experiments, features are standardized. Significance is then assessed at $p{<}0.05$.

\subsection{Logistic regression}
\label{app:meta-logit}

For each model we fit a logistic regression of correctness on the meta-features
(\texttt{scikit-learn} for coefficients and metrics; \texttt{statsmodels} Logit for $z$-statistics, $p$-values, and 95\% confidence intervals).
Cross-validated ROC-AUC (5-fold stratified) ranges from about 0.58 to 0.72 across models, so meta-features carry a moderate but incomplete signal.

Table~\ref{tab:meta-logit-summary} aggregates coefficients over the 13 Logit fits that converged (three models: Qwen3.5~397B, Qwen3-VL-2B-Instruct, and Qwen3-VL-8B-Instruct were excluded from the significance pool because of non-convergence, driven mainly by the rare \texttt{contains\_code} flag). Four predictors are significant for a large majority of converged models and enter with a negative mean coefficient: longer ground truth answers, scientific domain, Russian language, and presence of figures. \texttt{contains\_code} is never significant in the converged pool.

\begin{table}[htbp]
\centering
\small
\caption{Logistic regression summary across converged models ($n{=}13$).
Mean coefficient is the average standardized Logit coefficient; ``\% sig.'' is the share of
converged models with $p{<}0.05$.}
\label{tab:meta-logit-summary}
\begin{tabular}{lrrr}
\toprule
Feature & Mean coef. & Mean $|z|$ & \% sig. \\
\midrule
\texttt{answer\_length} & $-0.456$ & 4.53 & 92.3 \\
\texttt{domain\_science} & $-0.396$ & 4.68 & 84.6 \\
\texttt{lang\_ru} & $-0.353$ & 4.49 & 84.6 \\
\texttt{contains\_figures} & $-0.163$ & 2.95 & 84.6 \\
\texttt{question\_length} & $+0.109$ & 2.14 & 53.8 \\
\texttt{contains\_diagrams} & $-0.086$ & 1.68 & 46.2 \\
\texttt{complexity} & $-0.157$ & 1.95 & 38.5 \\
\texttt{contains\_equations} & $+0.028$ & 0.71 & 7.7 \\
\texttt{contains\_code} & $-0.015$ & 0.65 & 0.0 \\
\bottomrule
\end{tabular}
\end{table}

\subsection{SHAP importance}
\label{app:meta-shap}

\begin{figure}[t]
  \centering
  \includegraphics[width=\linewidth]{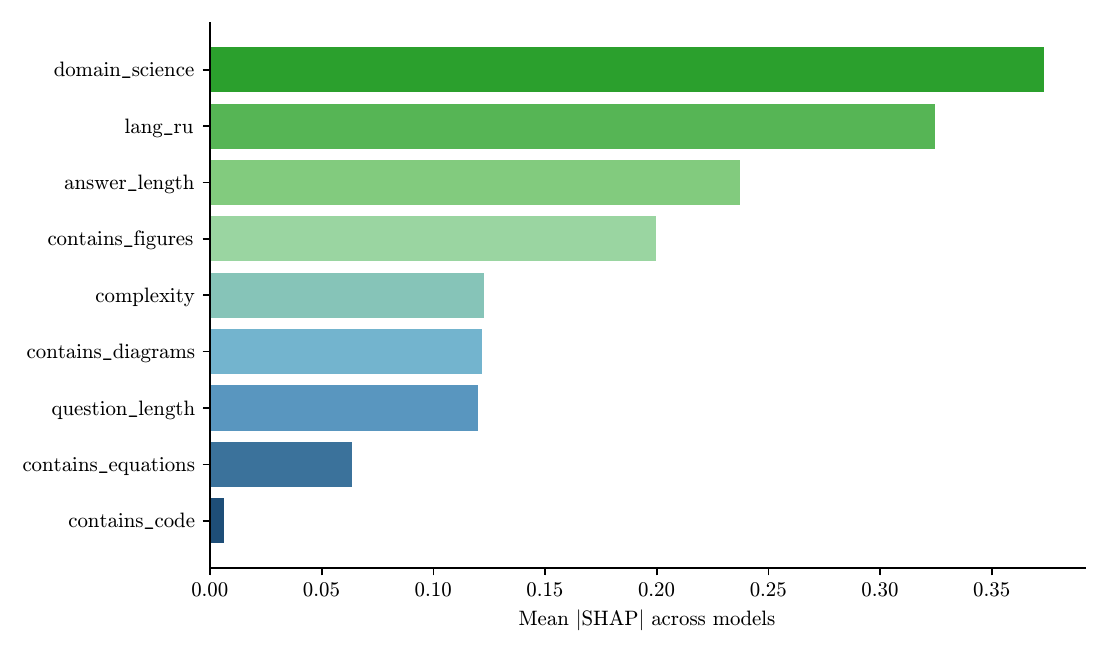}
  \caption{Mean absolute SHAP values of meta-features, averaged over 16 per-model logistic regressions.}
  \label{fig:meta-shap}
\end{figure}

We attribute the same standardized logistic models with \texttt{shap.LinearExplainer}
and rank features by mean absolute SHAP value, then average ranks/importances over all 16 models. Figure~\ref{fig:meta-shap} shows the resulting mean $|$SHAP$|$ scores.
Domain and language dominate, followed by gold-answer length and figures;
\texttt{contains\_code} is negligible, consistent with its $\approx$0.5\% prevalence.

\subsection{Correlations with correctness}
\label{app:meta-corr}

\begin{figure*}[t]
  \centering
  \includegraphics[width=\textwidth,height=0.78\textheight,keepaspectratio]{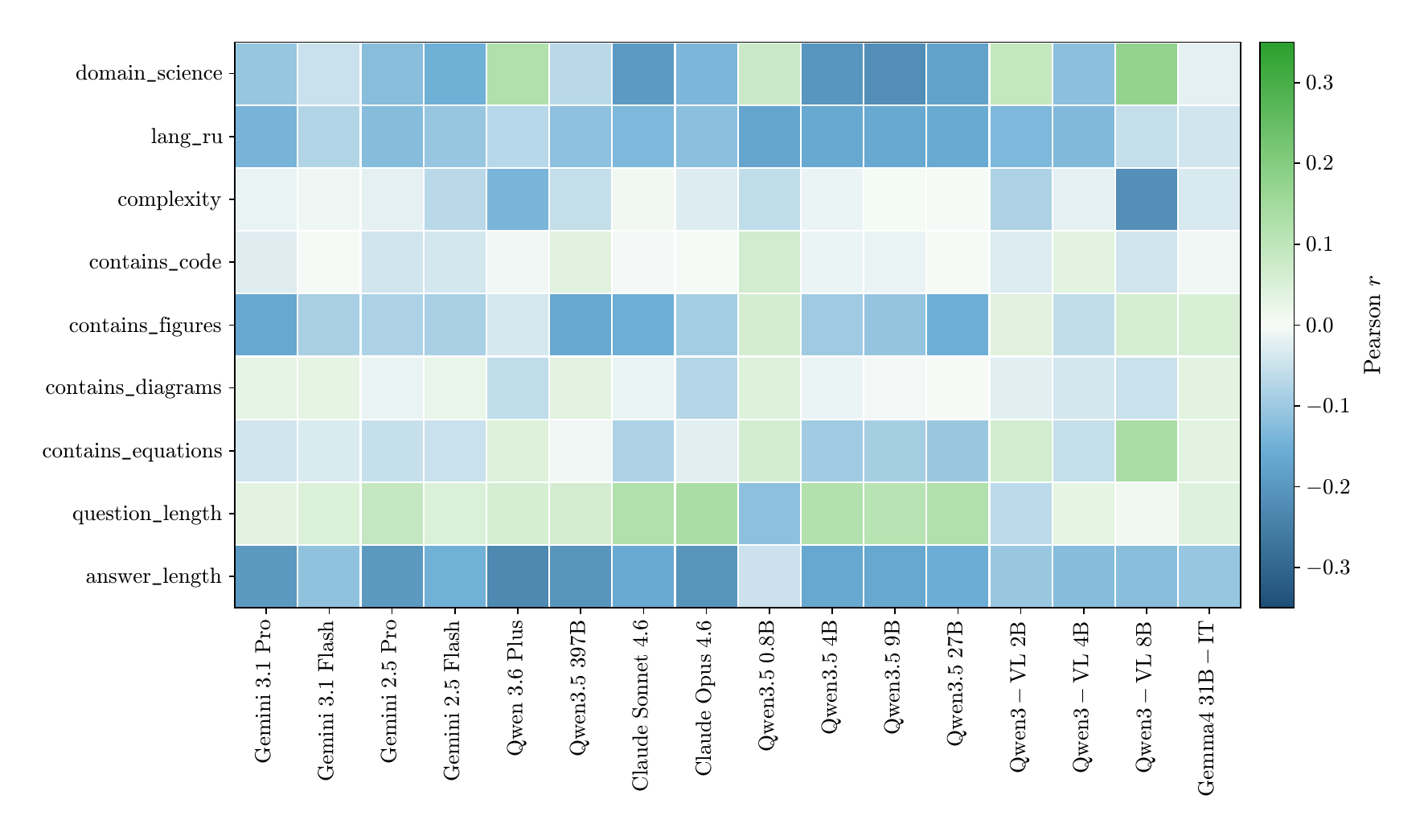}
  \caption{Per-model Pearson correlations between meta-features and label correctness}
  \label{fig:meta-corr}
\end{figure*}

Table~\ref{tab:meta-corr-summary} reports Pearson correlations between each meta-feature and binary correctness (equivalent to point-biserial correlation for binary predictors), averaged over models, together with the share of models where the two-sided test of $r{=}0$ is significant. Effect sizes are small ($|$mean $r$$|{\le}0.15$) but directionally stable: longer ground truth answers, Russian language, science domain, and items with figures tend to be harder.
Figure~\ref{fig:meta-corr} visualizes the full model$\times$feature correlation matrix.

\begin{table}[htbp]
\centering
\small
\caption{Pearson / point-biserial correlation of each meta-feature with correctness,
averaged over 16 models.}
\label{tab:meta-corr-summary}
\begin{tabular}{lrrr}
\toprule
Feature & Mean $r$ & Mean $|r|$ & \% sig. \\
\midrule
\texttt{answer\_length} & $-0.152$ & 0.152 & 93.8 \\
\texttt{domain\_science} & $-0.068$ & 0.127 & 87.5 \\
\texttt{lang\_ru} & $-0.119$ & 0.119 & 87.5 \\
\texttt{contains\_figures} & $-0.067$ & 0.093 & 62.5 \\
\texttt{question\_length} & $+0.054$ & 0.076 & 50.0 \\
\texttt{contains\_equations} & $-0.018$ & 0.061 & 43.8 \\
\texttt{complexity} & $-0.047$ & 0.048 & 25.0 \\
\texttt{contains\_diagrams} & $-0.006$ & 0.031 & 6.2 \\
\texttt{contains\_code} & $-0.005$ & 0.023 & 6.2 \\
\bottomrule
\end{tabular}
\end{table}

\subsection{Significance of features for models of different sizes}
\label{app:meta-size}

Finally we test whether differently sized models within the Qwen3.5 family rely on different meta-features. We compare Qwen3.5-0.8B, Qwen3.5-27B, and Qwen3.5-397B. \texttt{contains\_code} is dropped in this experiment to avoid quasi-complete separation.
For each size we fit a single Logit; for each size pair we stack item$\times$model rows and fit
\begin{equation*}
  \begin{aligned}
\mathrm{logit}\,P(\mathrm{correct})
&= X\beta + \gamma_0\,\mathbb{1}_{\mathrm{larger}} + \\
&\quad + (X\odot\mathbb{1}_{\mathrm{larger}})^\top\gamma,
\end{aligned}
\end{equation*}
with cluster-robust standard errors by item.
A significant interaction $\gamma_j$ means that feature $j$ has a different log-odds effect
in the larger model than in the smaller one.

Table~\ref{tab:meta-size-single} lists significant single-model coefficients. Table~\ref{tab:meta-size-delta} reports the interaction tests. The hypothesis is partially supported: 0.8B differs from both 27B and 397B on \texttt{domain\_science}, \texttt{contains\_figures}, and \texttt{question\_length} (and additionally on \texttt{contains\_equations} vs.\ 27B). The 27B vs.\ 397B contrast is much weaker, with only \texttt{lang\_ru} reaching $p{<}0.05$ (and only marginally). In short, the main gap is between the smallest model and the mid/large models rather than between 27B and 397B.

\begin{table}[htbp]
\centering
\scriptsize
\caption{Significant meta-features ($p{<}0.05$) in single Logit fits for three Qwen3.5 sizes.}
\label{tab:meta-size-single}
\begin{tabular}{lll}
\toprule
Size & Feature & Coef. \\
\midrule
\multirow{2}{*}{0.8B}
  & \texttt{lang\_ru}         & $-0.47$ \\
  & \texttt{question\_length} & $-0.40$ \\
\midrule
\multirow{6}{*}{27B}
  & \texttt{lang\_ru}          & $-0.49$ \\
  & \texttt{domain\_science}   & $-0.53$ \\
  & \texttt{contains\_figures} & $-0.35$ \\
  & \texttt{answer\_length}    & $-0.40$ \\
  & \texttt{question\_length}  & $+0.26$ \\
  & \texttt{contains\_diagrams}& $-0.17$ \\
\midrule
\multirow{5}{*}{397B}
  & \texttt{answer\_length}    & $-0.45$ \\
  & \texttt{contains\_figures} & $-0.37$ \\
  & \texttt{domain\_science}   & $-0.42$ \\
  & \texttt{lang\_ru}          & $-0.32$ \\
  & \texttt{complexity}        & $-0.27$ \\
\bottomrule
\end{tabular}
\end{table}

\begin{table*}[htbp]
\centering
\small
\caption{Tests of coefficient differences between Qwen3.5 sizes (interaction $\Delta$, larger $-$ smaller).
Only features with $p{<}0.05$ are shown.}
\label{tab:meta-size-delta}
\begin{tabular}{llrrr}
\toprule
Pair & Feature & $\Delta$ coef. & $z$ & $p$ \\
\midrule
0.8B vs.\ 27B
  & \texttt{question\_length} & $+0.660$ & 4.46 & $8.1{\times}10^{-6}$ \\
  & \texttt{contains\_figures} & $-0.534$ & $-4.14$ & $3.5{\times}10^{-5}$ \\
  & \texttt{domain\_science} & $-0.565$ & $-3.74$ & $1.9{\times}10^{-4}$ \\
  & \texttt{contains\_equations} & $-0.282$ & $-2.17$ & $0.030$ \\
\midrule
27B vs.\ 397B
  & \texttt{lang\_ru} & $+0.173$ & 1.96 & $0.050$ \\
\midrule
0.8B vs.\ 397B
  & \texttt{contains\_figures} & $-0.557$ & $-4.20$ & $2.7{\times}10^{-5}$ \\
  & \texttt{question\_length} & $+0.514$ & 3.32 & $9.2{\times}10^{-4}$ \\
  & \texttt{domain\_science} & $-0.460$ & $-2.92$ & $0.003$ \\
\bottomrule
\end{tabular}
\end{table*}

\paragraph{Takeaway.}
Across logistic regression, SHAP, and correlations, a consistent core of meta-features is associated with lower correctness: ground truth answer length, scientific domain, Russian language, and the presence of figures. In practice, this translates to a pronounced drop in answer quality when the prompt is in Russian, contains figures, or requires a long, detailed response. Furthermore, the comparison across model sizes reveals a distinct scaling effect: as the model size changes, its sensitivity to specific features shifts, indicating that models of different scales attend to different aspects of the input when generating answers. Within Qwen3.5, these behavioral shifts differ significantly between the 0.8B model and the larger variants.

\begin{figure*}[p]
  \centering
  \begin{subfigure}{\textwidth}
    \centering
    \includegraphics[width=\linewidth,height=0.42\textheight,keepaspectratio]%
      {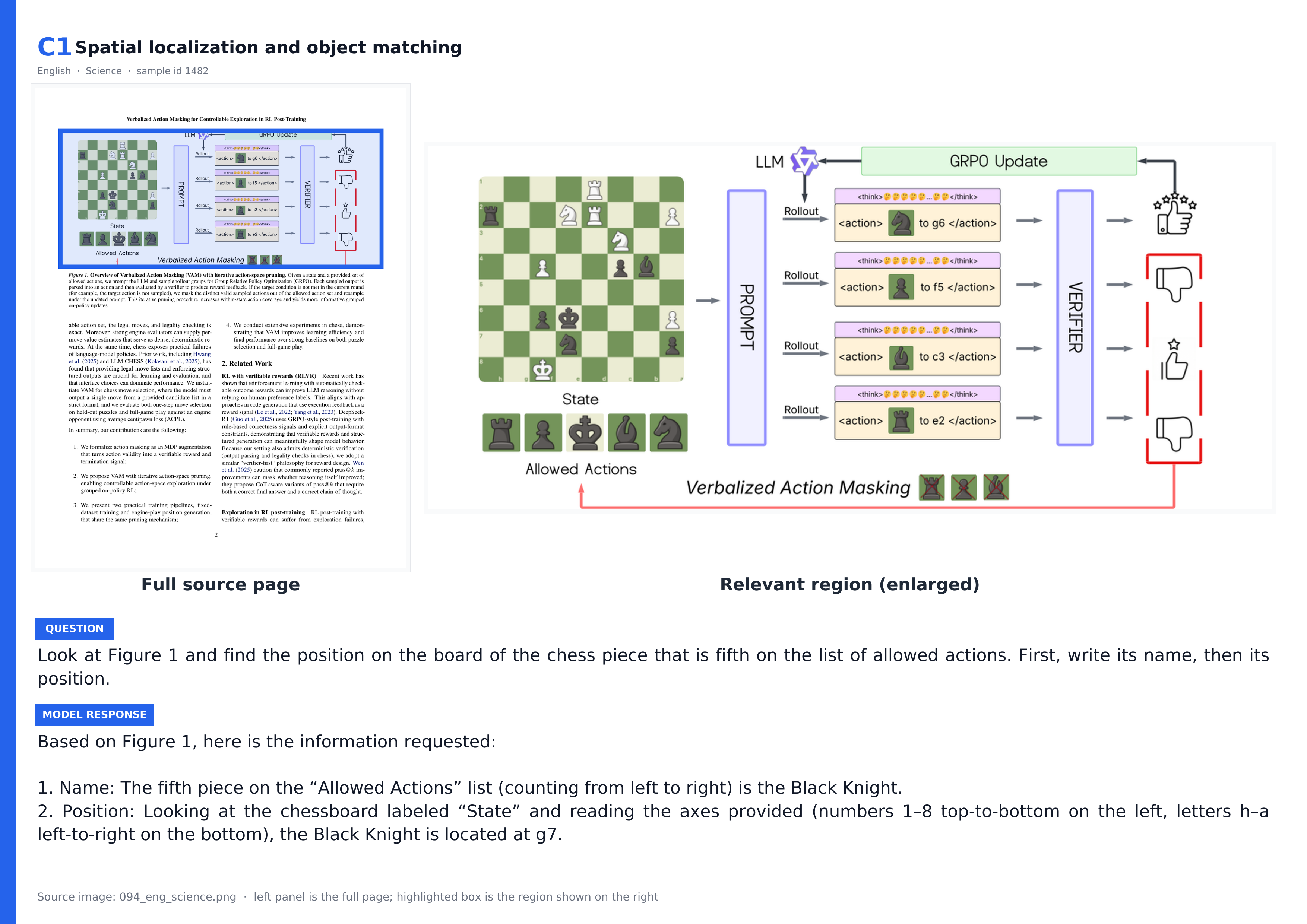}
    \caption{C1: spatial localization (EN, science).}
    \label{fig:err-c1}
  \end{subfigure}

  \vspace{0.8em}
  \begin{subfigure}{\textwidth}
    \centering
    \includegraphics[width=\linewidth,height=0.42\textheight,keepaspectratio]%
      {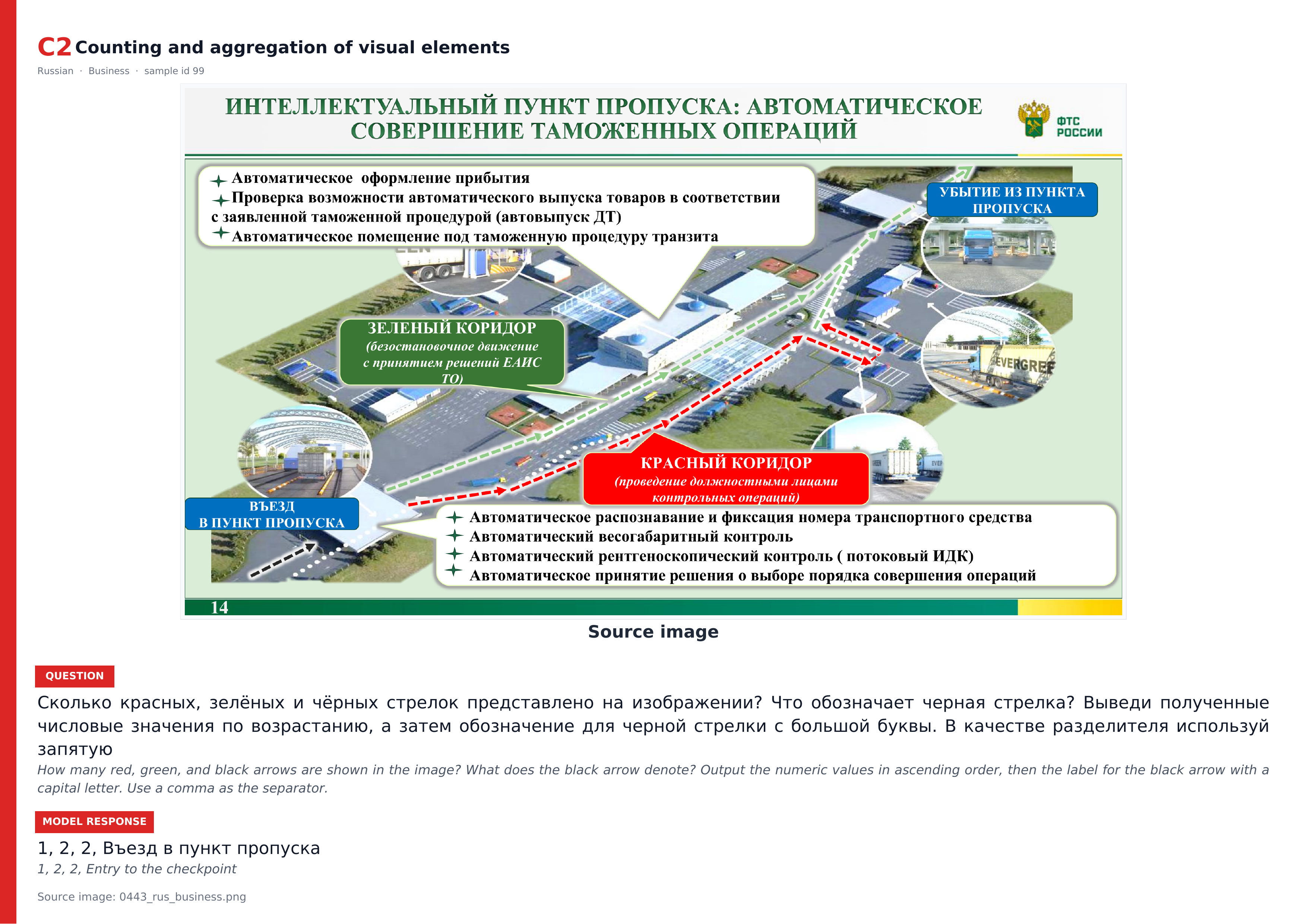}
    \caption{C2: counting (RU, business).}
    \label{fig:err-c2}
  \end{subfigure}

  \caption{Representative failures for error types C1--C5 (part~1 of~3).
    Each panel shows the source page, the question, and the model's answer.}
  \label{fig:error-types}
\end{figure*}

\begin{figure*}[p]
  \ContinuedFloat
  \centering
  \begin{subfigure}{\textwidth}
    \centering
    \includegraphics[width=\linewidth,height=0.42\textheight,keepaspectratio]%
      {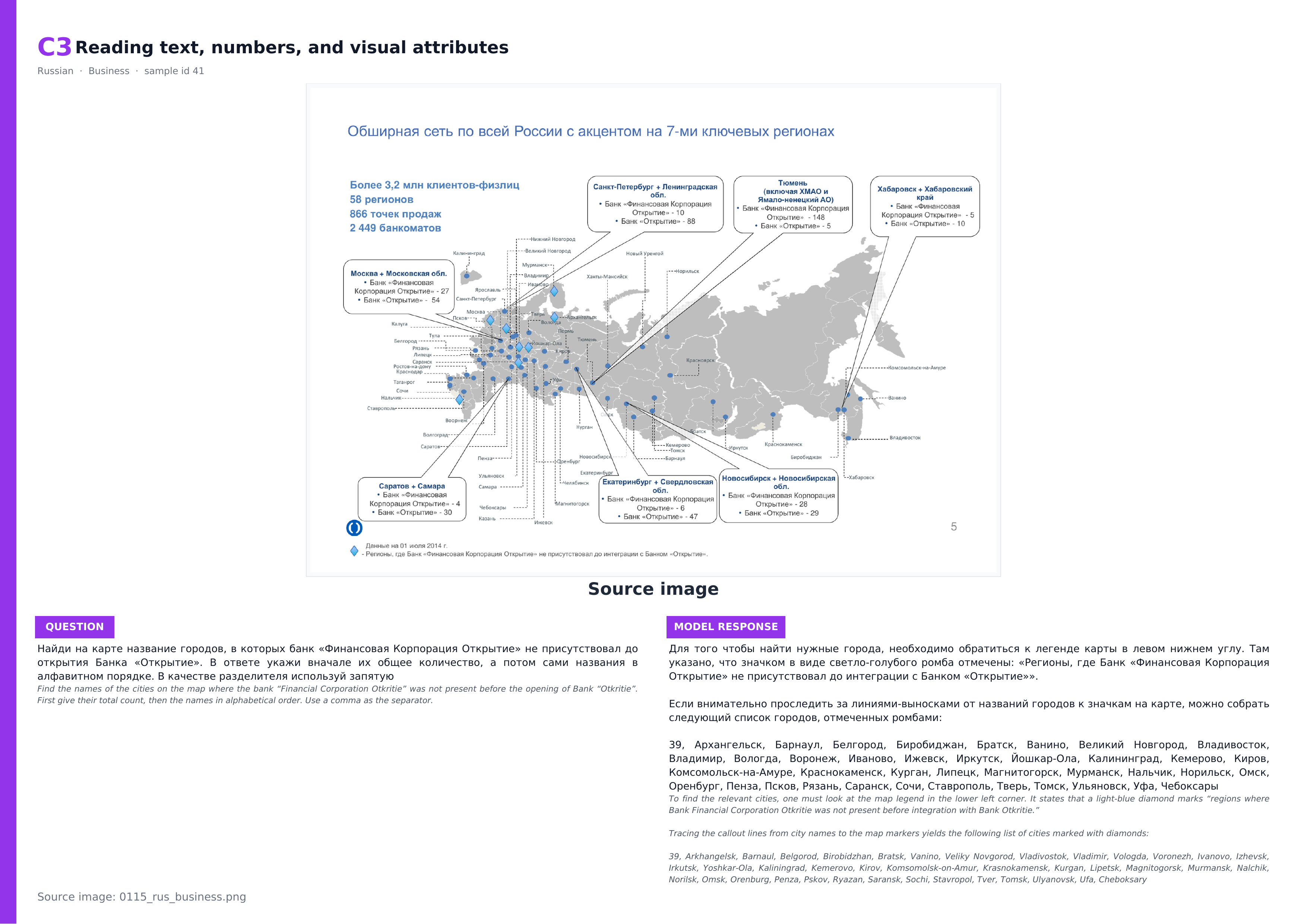}
    \caption{C3: visual attributes (RU, business).}
    \label{fig:err-c3}
  \end{subfigure}

  \vspace{0.8em}
  \begin{subfigure}{\textwidth}
    \centering
    \includegraphics[width=\linewidth,height=0.42\textheight,keepaspectratio]%
      {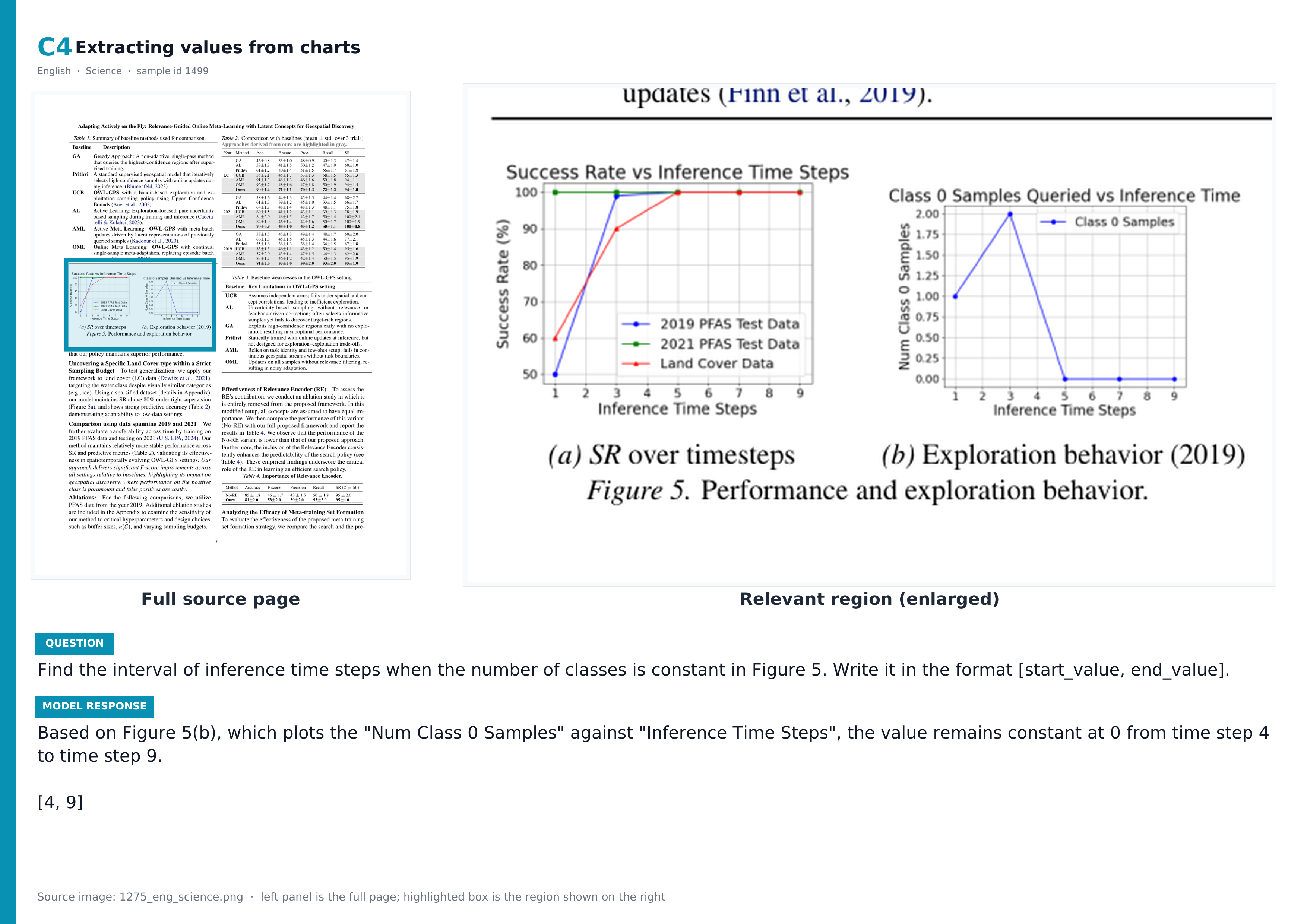}
    \caption{C4: chart reading (EN, business).}
    \label{fig:err-c4}
  \end{subfigure}

  \caption{Representative failures for error types C1--C5 (part~2 of~3).}
\end{figure*}

\begin{figure*}[p]
  \ContinuedFloat
  \centering
  \begin{subfigure}{\textwidth}
    \centering
    \includegraphics[width=\linewidth,height=0.80\textheight,keepaspectratio]%
      {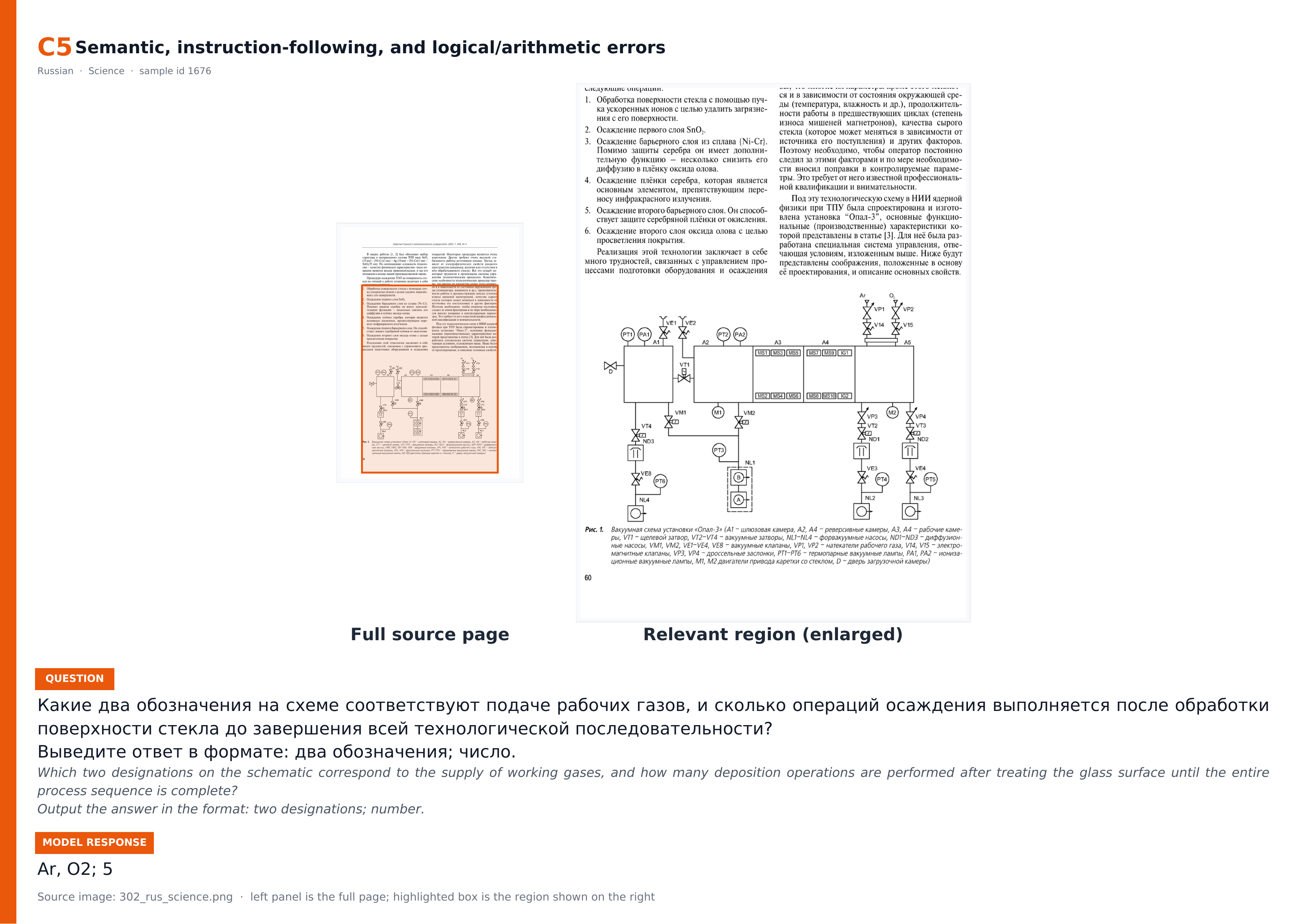}
    \caption{C5: semantic / instruction error (RU, science).}
    \label{fig:err-c5}
  \end{subfigure}

  \caption{Representative failures for error types C1--C5 (part~3 of~3).
    Examples cover English and Russian, and business and scientific documents.}
\end{figure*}
\end{document}

%% file: intro.tex
\section{Introduction}
\label{sec:intro}

Multimodal Large Language Models
(MLLMs)~\cite{openai2024gpt4technicalreport, geminiteam2025gemini, anthropic2024claude3} have revolutionized the
machine comprehension of images. Beyond merely solving optical
character recognition (OCR) tasks, these models demonstrate
the potential to perform complex reasoning over multimodal
inputs~\cite{yue2024mmmu,hao2025emma}. This capability is
particularly vital for text-rich scenarios, which are central
to real-world professional domains that require reading and analyzing dense
visual documents, such as scientific papers and financial
reports.

Existing benchmarks provide complementary but incomplete
coverage of professional document reasoning. Document-oriented
datasets such as DocVQA~\cite{mathew2021docvqa} primarily
emphasize OCR and information extraction, whereas broad
multimodal reasoning benchmarks such as
MMMU~\cite{yue2024mmmu} often require specialized factual
knowledge. OCR-Reasoning~\cite{huang2026ocrreasoning} covers
diverse text-rich images, including some professional
documents, but does not examine this setting in depth.
A further limitation is linguistic coverage: existing multimodal benchmarks exhibit a strong bias toward English and Chinese, leaving Slavic languages~---~and Russian in particular~---~substantially underrepresented.
Russian-inclusive resources such as
MTVQA~\cite{tang2025mtvqa}, MWS Vision
Bench~\cite{mwsvisionbench2025}, and
MERA-Multi~\cite{chervyakov2026meramulti} broaden this
coverage, but mix documents with other image types, emphasize
OCR and document processing, or focus on a narrow document
category. To our knowledge, no existing benchmark evaluates self-contained, multi-step reasoning over Russian-language professional document pages whose items involve textual and graphical content such as figures, tables, charts, equations, and diagrams.

To address these limitations, we introduce
\textbf{BEAR-Bench}, a complex benchmark comprising
1000 human-annotated questions for
text-dense enterprise and scientific documents in English and
Russian languages. The science section tests the ability to
interpret academic figures, transcribe
mathematical formulas, and analyze data from plots. The
enterprise tasks target data cross-referencing in
financial reports and logical reasoning over business charts.
The questions follow two design principles. First, they
require multiple logical steps rather than direct extraction
from a single piece of evidence. Second, they are fully
answerable from the document alone, without external expert
knowledge. Together, these principles focus the evaluation on
document-grounded reasoning rather than factual recall or
shallow extraction. The inclusion of Russian-language tasks
further broadens evaluation beyond predominantly
English- and Chinese-language resources.

Deploying MLLMs in professional settings requires knowing not
only how often they fail, but also whether those failures can
be detected reliably; yet evidence on hallucination detection
for reasoning over text-dense professional documents remains incomplete. We
address this gap by using BEAR-Bench to compare token-level
uncertainty scores, representation-based detectors,
supervised hidden-state probes, and MLLM-as-a-judge methods
across proprietary and open-weight models, providing a
practical comparison for deployment settings with and
without access to model internals.

The main contributions of this work are the following:
\begin{enumerate}
    \item We propose \textbf{BEAR-Bench}, a bilingual
    benchmark for multimodal reasoning in professional
    scenarios. It features human-annotated questions targeting
    text-dense business and scientific documents in English
    and Russian languages.
    \item We evaluate
    16 MLLMs on BEAR-Bench, including
    proprietary (e.g., Gemini-3.1-Pro, Claude Opus 4.6) and
    open-weight (e.g., Qwen3.5, gemma-4) models. Our
    results show that even the strongest evaluated systems
    leave clear headroom on BEAR-Bench.
    \item We further use BEAR-Bench to evaluate a diverse set
    of existing hallucination detection methods for
    OCR-intensive professional-document reasoning,
    comparing uncertainty-, representation-, and
    judge-based methods under two deployment regimes: direct
    access to internal signals for open-weight models and
    proxy-based detection for proprietary ones.
\end{enumerate}

\begin{table*}[t!]
\centering
\small
\resizebox{\textwidth}{!}{%
\begin{tabular}{lccclc}
\toprule
\textbf{Benchmark} & \textbf{\#Langs} & \textbf{\#QA pairs} & \textbf{\#RU reasoning VQA} & \textbf{Image scope} & \textbf{OCR chars/img} \\
\midrule
DocVQA~\cite{mathew2021docvqa}         & EN & 5.2K   & \na & Industry documents & 1{,}113.0 \\
ChartQA~\cite{masry2022chartqa}         & EN & 2.5K*  & \na & Charts & 231.8 \\
CharXiv~\cite{wang2024charxiv}         & EN & 11.6K & \na & Scientific charts & 165.1 \\
OCRBench v2~\cite{fu2025ocrbenchv2}     & EN, ZH & 10K    & \na & Mixed text-rich images & 437.2 \\
OCR-Reasoning~\cite{huang2026ocrreasoning}  & EN & 1.1K    & \na & Everyday text-rich scenes & 514.3 \\
\midrule
MWS Vision Bench~\cite{mwsvisionbench2025}
                   & RU & 1.3K**   & 400 & Business/personal documents & 1{,}126.4 \\
LabTabVQA~\cite{chervyakov2026meramulti}
                   & RU & 349   & 349 & Medical report tables & 637.1 \\
CC-OCR V2~\cite{xu2026ccocr} & RU + 31 langs & 2K*** & 0 & Finance/dashboards/blueprints &  1{,}149.6\\
\midrule
BEAR-Bench  & EN, RU & 1K & 618 & Science/business documents & 2{,}740.6 \\
\bottomrule
\end{tabular}}
\caption{Comparison of BEAR-Bench to existing multimodal reasoning benchmarks. \textbf{\#QA pairs} refers to the evaluation split; DocVQA and ChartQA additionally provide training data (50K and 32.7K items in total, respectively), while the remaining benchmarks are evaluation-only. For CC-OCR V2, the reported count is the document QA track (2K of 7.1K items). \textbf{\#RU reasoning VQA}: number of evaluation questions in Russian that require answering from the image, excluding OCR, parsing, grounding, and key-information extraction; n/a~=~not applicable (no Russian split). \textbf{Image scope} summarizes the principal visual sources represented in each benchmark. \textbf{OCR chars/img}: average number of non-whitespace OCR characters per image ($n{=}200$ randomly sampled images per benchmark). \textit{*Evenly split between human-written and machine-generated questions. **Publicly available.}.}
\label{tab:comparison}
\end{table*}

%% file: related_work.tex
\section{Related Work}
\label{sec:related}
\paragraph{Multimodal benchmarks for professional documents.} 
While modern LLMs achieve strong results on many established multimodal benchmarks~\cite{yue2025mmmu, zuo2025medxpertqa}, their ability to operate with visually rich professional documents~---~which requires analyzing both textual and graphical content of an image~----~remains underexplored. Text-dense benchmarks oriented at professional documents are mostly OCR-based and measure extractive skills rather than cross-referencing of text and visuals~\cite{xu2026ccocr, mathew2021docvqa}, or emphasize long-page settings that conflate multimodal reasoning with long-context handling~\cite{chen2026scimdr, tang2026finmmdocr}. Multimodal reasoning-focused benchmarks, conversely, offer little signal specifically for professional documents: OCR-Reasoning~\cite{huang2026ocrreasoning} and OCRBench v2~\cite{fu2025ocrbenchv2} are deliberately broad~---~valuable for general-purpose evaluation, but treating professional documents as one setting among many~---~while CharXiv~\cite{wang2024charxiv} and ChartQA~\cite{masry2022chartqa, masry2025chartqapro} restrict evaluation to charts and thus do not test text–visual cross-referencing. A further issue is dependence on external knowledge: MMMU~\cite{yue2024mmmu} and EMMA~\cite{hao2025emma} pose multidisciplinary problems that presuppose domain expertise, so scores conflate multimodal reasoning failures with factual gaps; part of OCR-Reasoning shares this confound.

Language coverage is also narrow: most benchmarks covering multimodal reasoning in text-dense scenarios are
available only in English or Chinese, while many other languages, including Russian, remain underrepresented. Russian-inclusive
benchmarks provide valuable but
partial coverage. MTVQA~\cite{tang2025mtvqa} and TIU-Bench~\cite{zhang-etal-2025-tiu} include documents alongside natural scenes but offer very limited coverage of Russian-language professional documents; TIU-Bench, for example, contains only 10 Russian document samples in total. MWS Vision
Bench~\cite{mwsvisionbench2025} provides 400 Russian
reasoning-VQA items, but its images span business scans,
personal handwriting, receipts, and form-style pages
(Figure~\ref{fig:mws-examples}), rather than being selected
specifically for reasoning over text-dense professional
documents. \textit{LabTabVQA}~---~the only
subset of MERA-Multi~\cite{chervyakov2026meramulti} built on
professional documents rather than natural images or
exam-style problems~---~is restricted to tables from
medical laboratory reports. 

\paragraph{Error detection for text-dense multimodal inputs.}
Recent work proposes a range of hallucination detectors for MLLMs~\cite{chen2026survey}. Tool-augmented methods rely on auxiliary models~\cite{chen2024unified,yin2024woodpecker,sahu-etal-2024-pelican}, while multi-query methods require repeated generation or verification calls~\cite{wu2024logical,zhang2024vl}, increasing deployment cost. Of particular interest are lightweight white-box methods, which detect errors using token uncertainty or internal model states collected during a single forward pass~\cite{tong2026faithscan,li2024reference,jiang2025devils,zhang2026vib}. These methods add relatively little inference overhead when model internals are available, yet, to our knowledge, they have not been systematically compared on visually rich professional document images requiring multi-step reasoning across dense textual and graphical evidence.

\paragraph{Summary.} Existing work lacks a Russian-inclusive
benchmark for self-contained, multi-step reasoning over
text-dense professional documents. BEAR-Bench is designed to fill this
gap using scientific and business documents;
Table~\ref{tab:comparison} summarizes the comparison.

Furthermore, BEAR-Bench enables a systematic comparison of hallucination detectors on such text-dense professional document images requiring multi-step reasoning, a setting not covered by prior evaluations.

%% file: benchmark.tex
\section{BEAR-Bench}
\label{sec:benchmark}

\subsection{Domain Scope and Taxonomy}

BEAR-Bench spans two primary domains~---~\textbf{Business} and
\textbf{Science}~---~each subdivided into thematically coherent
sub-categories.

\paragraph{Business Domain.}
The business subset covers three document types:
\textit{financial reports} (SEC Forms~10-K and~10-Q)
requiring tabular reasoning and year-over-year calculations;
\textit{investor presentations} (Form~8-K exhibits)
combining charts, KPI tiles, and infographic maps; and
\textit{flowcharts and organisational diagrams} depicting
corporate ownership structures and process pipelines.

\paragraph{Science Domain.}
The science subset covers three categories:
\textit{mathematical and physical formulae} from physics and
mathematics preprints targeting symbol-level recognition;
\textit{scientific figures and plots} (line plots, scatter
diagrams, heatmaps) requiring axis and legend
interpretation; and \textit{academic layouts} with
multi-column pages testing reading-order resolution and
cross-referential reasoning.

The two domains are strictly disjoint: no source document
appears in both subsets.

\subsection{Data Collection and Annotation Pipeline}

\subsubsection{Source Collection}

\paragraph{Business Domain.}
Business documents were retrieved via targeted Google Search
queries directed at publicly accessible, license-safe
sources. English-language documents were obtained from the
U.S.\ Securities and Exchange Commission (SEC) EDGAR
system~---~annual reports (Form~10-K), quarterly reports
(Form~10-Q), and investor presentations filed as Form~8-K
exhibits~---~all of which constitute public records under
U.S.\ federal law. Supplementary English documents were
drawn from official government portals (\texttt{*.gov},
\texttt{*.gov.uk}) and intergovernmental repositories
(\texttt{*.int}). Russian-language documents were sourced
from the state corporate-disclosure platforms
\texttt{e-disclosure.ru} and \texttt{moex.com}, as well as
from federal government domains (\texttt{*.gov.ru}). An
automated scraper retrieved candidate PDFs; each document
underwent a programmatic license-verification step
examining the first and last five pages for SEC registration
markers or open-license declarations
(\emph{``Creative Commons''}, \emph{``CC~BY''},
\emph{``public domain''}). Documents failing this check
were discarded prior to further processing.

\paragraph{Science Domain.}
English-language papers were downloaded from arXiv via its
official Python API, sampling four STEM categories:
\texttt{quant-ph}, \texttt{cs.AI}, \texttt{eess.SP}, and
\texttt{math.GM} (up to 100 papers per category).
Russian-language articles were collected from CyberLeninka
(\texttt{cyberleninka.ru}) using an asynchronous
Playwright-based crawler across four subject areas:
Computer Science, Mathematics, Physics, and Engineering
(up to 100 articles per category).

\subsubsection{Filtering and Preprocessing}

Raw PDFs were rendered page-by-page into PNG images and
processed through a two-stage filtering pipeline.

\paragraph{Stage~1\,---\,Visual Content Classification.}
We obtained silver labels for a stratified sample of
3{,}000 images using Gemini~2.5~Pro with a structured
multi-label prompt, producing six Boolean fields:
\texttt{contains\_diagrams}, \texttt{contains\_tables},
\texttt{contains\_equations}, \texttt{contains\_code},
\texttt{contains\_figures}, and
\texttt{contains\_handwriting}. These labels trained a
lightweight classifier: SigLIP
embeddings~\citep{zhai2023sigmoid} were L2-normalised and
passed to a \texttt{MultiOutputClassifier} of
logistic-regression models with balanced class weights,
one per label. After validation on a held-out 20\% split,
the classifier was applied to the full corpus of
$\sim$66{,}000 page images, retaining only pages with at
least one of \{\texttt{contains\_diagrams},
\texttt{contains\_equations}, \texttt{contains\_code}\}
predicted positive.

\paragraph{Stage~2\,---\,Textual Density Filtering.}
Among content-positive pages, we retained only those at or
above the 67th percentile of OCR character count within
their respective language group. From the surviving
candidates, up to 1{,}500 images per language were drawn
via stratified random sampling (seed\,${=}\,42$), yielding
the final pool submitted to human annotators.

\begin{figure*}[t!]
  \centering
  \includegraphics[width=\textwidth,height=0.28\textheight,keepaspectratio]{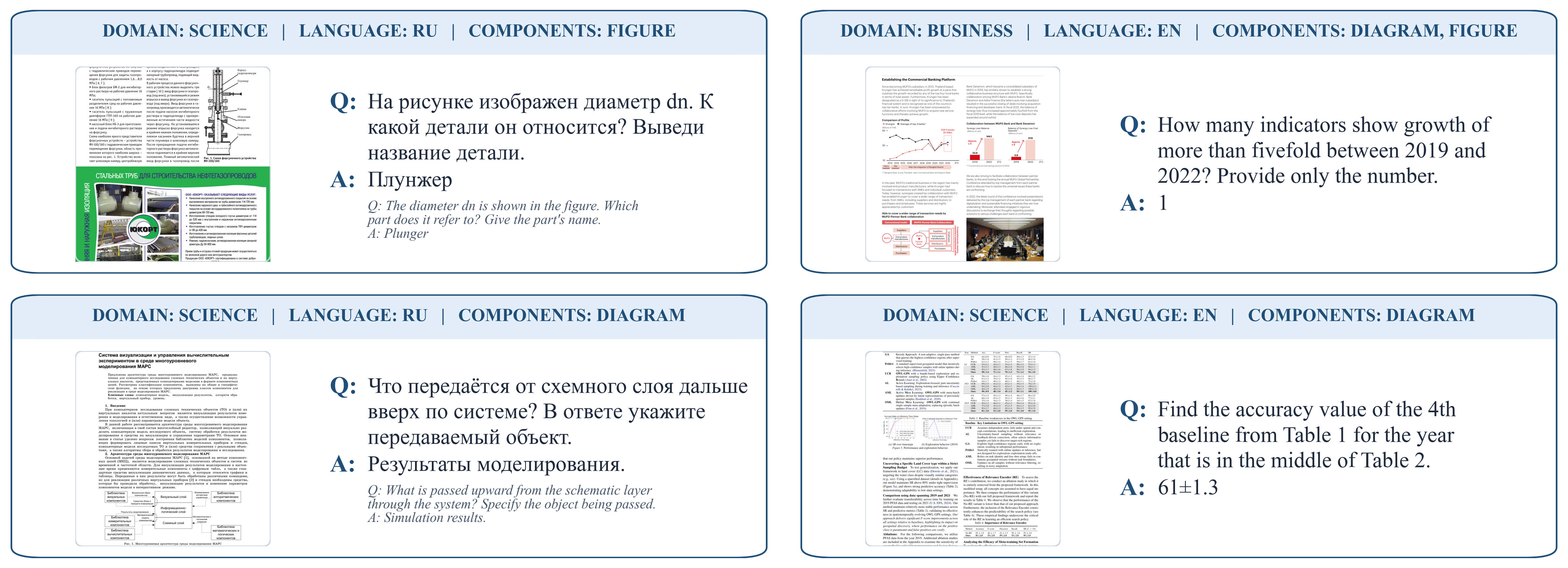}
  \caption{%
    \textbf{Representative items from BEAR-Bench.}
    Each card displays the source document image (left)
    alongside the human-authored multi-hop question and
    ground-truth answer (right).%
  }
  \label{fig:examples}
\end{figure*}

\subsubsection{Human Annotation}

\paragraph{Annotator pool.}
Annotation was conducted by 13 domain experts, each holding
at minimum a Bachelor's degree in a technical discipline.
Every annotator processed its own subset of images, authoring exactly one question--answer pair
per image.

\paragraph{Annotation task.}
For each image, annotators were required to:
(i)~re-verify the visual-content labels produced by the
automatic classifier, correcting any erroneous predictions;
and (ii)~compose a multi-step, multi-hop question with a
detailed ground-truth answer. Questions were required to elicit compositional
reasoning~---~aggregating values across table rows,
interpreting plotted trends in the context of equations, or
tracing paths through flowcharts~---~rather than
straightforward single-step extraction. The questions were
additionally assigned a \textit{reasoning depth score}
representing the total number of reasoning and computational
steps required to arrive at the correct answer. The text of the instruction for the annotators is reported in Figure~\ref{fig:annotator-instructions-ru}.

\paragraph{Evaluation judge.}
Model responses are scored by GPT-4o used as an
LLM-as-a-judge. The judge assesses semantic equivalence
between the model answer and the ground truth, permitting
surface-level paraphrase while penalising under-specific
responses. It returns a binary verdict $v \in
\{\texttt{true},\,\texttt{false}\}$ of whether the model answer is correct with a brief
explanation in structured XML tags, enabling fully
reproducible programmatic evaluation. The exact prompt is provided in Figure~\ref{fig:llm-judge-prompt}.

\paragraph{Judge reliability.}
To validate the reliability of the LLM-as-a-judge protocol, we
constructed a stratified audit sample of 200 judge verdicts,
drawn uniformly across languages and domains (100 English and
100 Russian items; 100 Business and 100 Science items, with
50 items per language--domain cell). Human annotators
independently reviewed each model response, ground-truth
answer, and judge verdict, recording agreement or
disagreement. The judge achieved an overall human-agreement
rate of 99.0\% (198/200), with only two
disagreements in the entire sample. Agreement remained
consistently high across languages (99\% for both
English and Russian) and domains (99\% for both
Business and Science), as well as at the finer-grained
language-domain level (98-100\% across all four
cells), with 95\% Wilson confidence intervals overlapping the
overall estimate throughout. On this stratified sample, the LLM-as-a-judge protocol agrees closely with human evaluation.

\paragraph{Quality control.}
Eight state-of-the-art proprietary VLMs were queried on
every item: Gemini~2.5~Pro/Flash, Gemini~3.1~Pro/Flash,
Qwen~3.6~Plus, Qwen3.5~397B, Claude~Sonnet~4.6, and
Claude~Opus~4.6. Items where three or more models returned
identical responses~---~normalised for punctuation and
case~---~and the LLM judge assigned \texttt{false} to all
answers were flagged. A manual audit confirmed that 99\% of
flagged items had erroneous or ambiguous ground truth; all
were excluded from the final benchmark. A random sample of retained items was quality-assessed along four dimensions: GT quality (84.4\%), judge verdict quality (90.6\%), question quality (90.9\%), and image quality
(97.0\%).

\paragraph{Final dataset composition.}
After quality-control filtering, BEAR-Bench comprises
\textbf{1{,}000 document images} paired with
\textbf{1{,}000 human-authored QA instances} across four
domain--language groups. Representative samples from BEAR-Bench are shown in Figure~\ref{fig:examples}. 

%% file: statistics.tex
\section{Dataset Statistics and Analysis}
\label{sec:statistics}

\begin{figure*}[t]
  \centering
  \includegraphics[width=\textwidth,height=0.24\textheight,keepaspectratio]{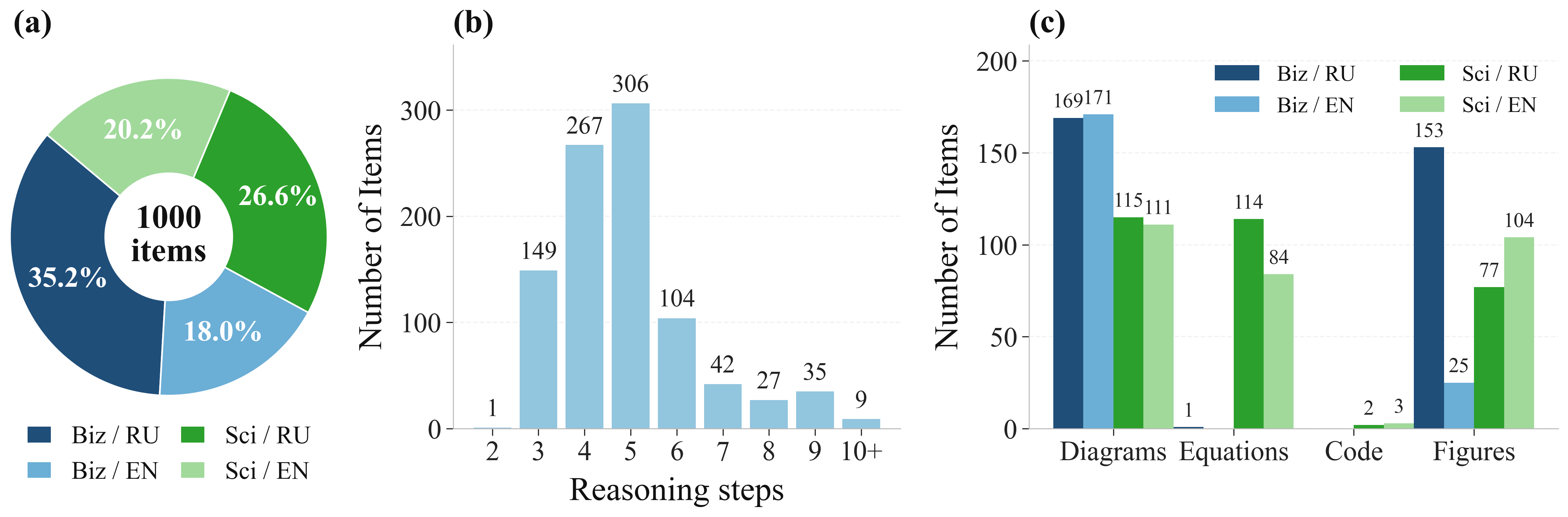}
  \caption{%
    \textbf{BEAR-Bench dataset statistics.}
    \textbf{(a)}~Item distribution across the four
    domain\,$\times$\,language cells; the central numeral
    indicates the total count.
    \textbf{(b)}~Distribution of complexity scores over all
    1{,}000 items, where each score reflects the total
    number of reasoning and computational steps required to
    solve the corresponding question.
    \textbf{(c)}~Prevalence of visual content
    types, diagrams, equations, code, and
    figures, disaggregated by subset; bars show absolute
    counts. Items may carry multiple content-type labels
    simultaneously.
  }
  \label{fig:dataset_stats}
\end{figure*}

Figure~\ref{fig:dataset_stats} summarises the composition
of BEAR-Bench across three dimensions: domain--language
balance, question complexity, and visual content-type
prevalence.

\paragraph{Domain and language balance.}
The Russian business cell is the largest subset
(352 items, 35.2\%), reflecting the higher volume of
publicly accessible Russian-language corporate disclosure
documents, while the English business cell is the smallest
(180 items, 18.0\%). The science cells are more evenly
distributed (266 and 202 items for Russian and English,
respectively).

\paragraph{Reasoning depth.}
Reasoning-depth annotations are available for 940 of the 1,000 items in BEAR-Bench. Each annotation estimates the intended number of steps required to derive the correct answer from the document image. Because such step counts depend on how annotators decompose a task, we treat them as coarse descriptive metadata rather than an objective difficulty score. The estimates span 2--10+ steps and peak at 4--5 steps with a moderate positive skew, indicating that the benchmark construction targeted multi-step inference rather than simple extraction. %

\paragraph{Visual content types.}
Figures and diagrams are the most prevalent content types
across all subsets, consistent with the heavy use of
infographics in both corporate reports and scientific
papers. Equations appear almost exclusively in the science
subsets, reflecting the mathematical nature of the arXiv
and CyberLeninka source material, while code fragments are
comparatively rare overall.

%% file: experiments.tex
\section{Experiments}
\label{sec:experiments}

\subsection{Experimental Setup}

\paragraph{Evaluated models.}

We evaluate a diverse set of MLLMs on BEAR-Bench, covering both open-weight models (Qwen3.5-0.8B/4B/9B/27B~\cite{qwen3.5}, Qwen3-VL-2B/8B-Instruct~\cite{qwen3technicalreport}, Qwen3-VL-4B-Thinking~\cite{qwen3technicalreport}, and Gemma-4-31B-it~\cite{gemmateam2026gemma4}) and proprietary systems (Qwen3.5-397B-A17B~\cite{qwen3.5}, Qwen 3.6 Plus~\cite{qwen36plus}, Gemini 2.5/3.1 Pro/Flash~\cite{geminimodelcards}, and Claude Sonnet/Opus 4.6~\cite{ClaudeSonnet46,ClaudeOpus46}). The open-weight models were run locally on an internal GPU server equipped with NVIDIA H100 and NVIDIA L40 accelerators; the proprietary ones were queried via the OpenRouter API.

\paragraph{Inference protocol.}
All models were evaluated zero-shot under a fixed protocol: each instance received only the image and the raw question, with no few-shot examples or prompt engineering, and a uniform decoding temperature of 0.6.

\subsection{Results}

\begin{table*}[t]

\centering
\caption{BEAR-Bench leaderboard. Results are reported as
  accuracy (\%) across all evaluation subsets.
  \textbf{Bold} denotes the best result in each column.}
\label{tab:leaderboard}
\resizebox{\textwidth}{!}{%
\begin{tabular}{@{}l r r r r r r r r@{}}
\toprule
\textbf{Model}
  & \textbf{Overall}
  & \textbf{Business}
  & \textbf{Science}
  & \textbf{RU}
  & \textbf{EN}
  & \textbf{Figures}
  & \textbf{Diagrams}
  & \textbf{Equations} \\
  & \small($n{=}1000$)
  & \small($n{=}532$)
  & \small($n{=}468$)
  & \small($n{=}618$)
  & \small($n{=}382$)
  & \small($n{=}359$)
  & \small($n{=}566$)
  & \small($n{=}199$) \\
\midrule
\multicolumn{9}{l}{\textit{Proprietary models}} \\
\midrule
Qwen3.5-397B-A17B
  & \textbf{75.4} & 78.2          & \textbf{72.2}
  & \textbf{71.4}         & 81.9 & \textbf{65.7}
  & \textbf{76.9} & \textbf{74.9} \\
Gemini 3.1 Pro
  & 75.1          & \textbf{79.5} & 70.1
  & 70.2          & \textbf{83.0}          & 65.5
  & 76.3          & 71.4 \\
Gemini 2.5 Pro
  & 67.4          & 72.7          & 61.3
  & 62.8          & 74.9          & 62.4
  & 66.8          & 62.3 \\
Qwen 3.6 Plus
  & 65.0          & 59.4          & 71.4
  & 62.5          & 69.1          & 62.7
  & 62.4          & 69.3 \\
Claude Sonnet 4.6
  & 62.7          & 71.4          & 52.8
  & 57.6          & 70.9          & 52.9
  & 62.2          & 54.8 \\
Claude Opus 4.6
  & 61.5          & 67.5          & 54.7
  & 57.1          & 68.6          & 55.7
  & 58.3          & 59.3 \\
Gemini 3.1 Flash
  & 59.8          & 62.2          & 57.0
  & 56.8          & 64.7          & 54.0
  & 61.3          & 56.8 \\
Gemini 2.5 Flash
  & 49.0          & 56.0          & 41.0
  & 44.8          & 55.8          & 43.2
  & 50.0          & 43.7 \\
\midrule
\multicolumn{9}{l}{\textit{Open-weight models}} \\
\midrule
Qwen3.5-27B
  & 60.8          & 68.9          & 51.6
  & 54.6          & 70.8          & 50.8
  & 60.8          & 50.8 \\
Qwen3.5-9B
  & 49.6          & 59.8          & 38.1
  & 43.3          & 60.0          & 42.5
  & 49.5          & 40.7 \\
Qwen3.5-4B
  & 48.6          & 58.1          & 37.9
  & 42.3          & 59.0          & 42.2
  & 48.0          & 39.2 \\
Qwen3-VL-4B-Thinking
  & 34.5          & 39.8          & 28.5
  & 29.7          & 42.4          & 30.7
  & 32.8          & 29.1 \\
Qwen3-VL-8B-Instruct
  & 25.4          & 18.3          & 33.4
  & 23.5          & 28.4          & 28.8
  & 23.4          & 37.2 \\
gemma-4-31B-it
  & 23.5          & 24.3          & 22.5
  & 22.0          & 25.8          & 26.5
  & 24.8          & 26.6 \\
Qwen3-VL-2B-Instruct
  & 14.3          & 11.3          & 17.8
  & 10.7          & 20.3          & 16.2
  & 13.7          & 19.1 \\
Qwen3.5-0.8B
  & 10.6          & 8.3           & 13.3
  & 6.5           & 17.4          & 13.1
  & 11.9          & 14.6 \\
\bottomrule
\end{tabular}}
\end{table*}

\begin{figure}[t]
    \centering
    \includegraphics[width=\columnwidth]{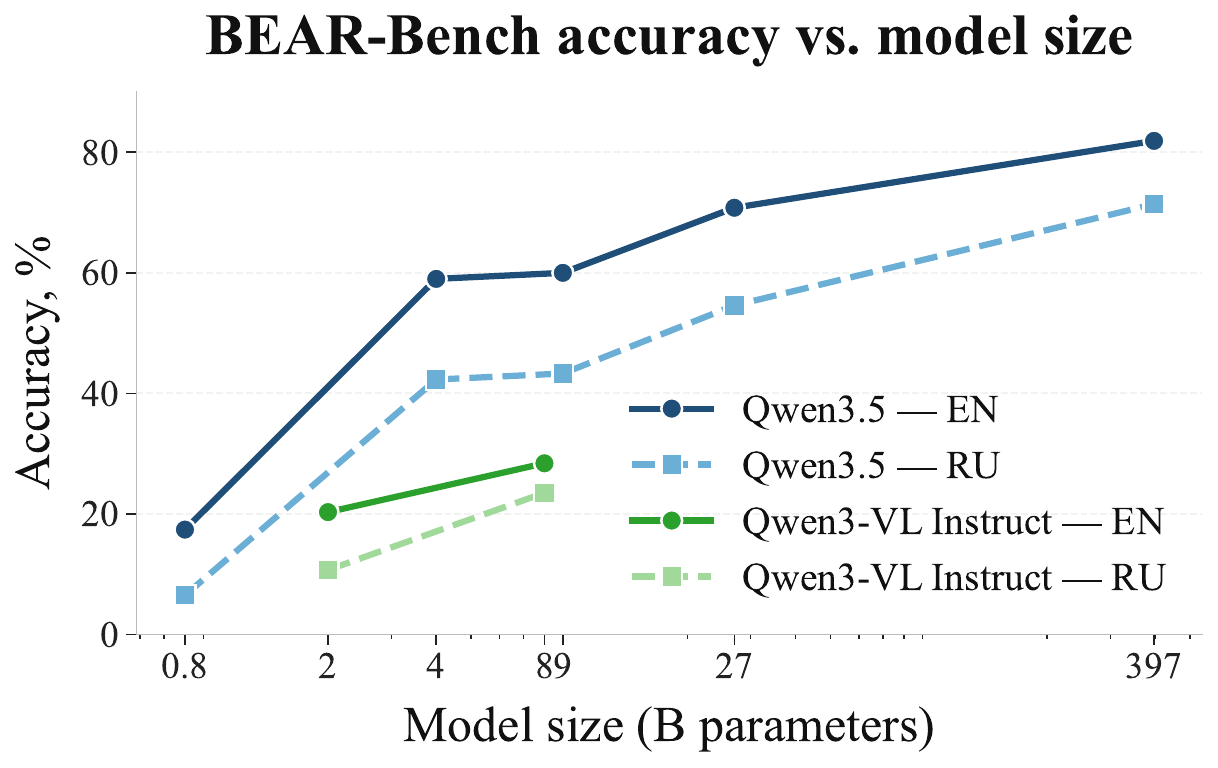}
    \caption{Accuracy on BEAR-Bench versus model size for the Qwen3.5 and Qwen3-VL-Instruct families (parameters on a log axis), reported separately for English (solid) and Russian (dashed) items. The English-over-Russian gap persists across scales.}
    \label{fig:qwen_scaling}
\end{figure}

\paragraph{Main results.} Table~\ref{tab:leaderboard} shows remaining headroom on BEAR-Bench. Qwen3.5-397B-A17B and Gemini~3.1~Pro achieve the highest overall accuracy (75.4\% and 75.1\%, respectively), trading the lead across subsets~---~Qwen3.5-397B-A17B is stronger on Science and Equations, while Gemini~3.1~Pro edges ahead on Business and English items~---~indicating that no single system dominates across all domains. All models exhibit a marked drop from English to Russian (e.g., 83.0\%$\rightarrow$70.2\% for Gemini~3.1~Pro and 81.9\%$\rightarrow$71.4\% for Qwen3.5-397B-A17B), confirming that the linguistic gap identified in prior benchmarks persists even for frontier proprietary models. Among the evaluated systems, proprietary models generally achieve higher accuracy than their open-weight counterparts. Within the Qwen3.5 and Qwen3-VL-Instruct families, larger models tend to perform better in both languages (Figure~\ref{fig:qwen_scaling}), although accuracy remains substantially lower on Russian items. Overall, current models remain limited in visually grounded reasoning over text-dense professional documents, with performance shaped jointly by model scale, language, and document type.

\paragraph{Effect of Chain-of-Thought prompting.}

\begin{table}[t]
\centering
\caption{Overall accuracy (\%) with and without an explicit
  chain-of-thought prompt. $\Delta$ is CoT minus the default
  (no-CoT) protocol used in Table~\ref{tab:leaderboard}.}
\label{tab:cot}
\small
\begin{tabular}{@{}l r r r@{}}
\toprule
\textbf{Model}
  & \textbf{No CoT}
  & \textbf{CoT}
  & $\Delta$ \\
\midrule
Qwen3.5-9B
  & 49.6 & 49.3 & $-$0.3 \\
Qwen3.5-4B
  & 48.6 & 47.2 & $-$1.4 \\
Qwen3-VL-4B-Thinking
  & 34.5 & 34.9 & $+$0.4 \\
Qwen3-VL-8B-Instruct
  & 25.4 & 39.1 & $+$13.7 \\
\bottomrule
\end{tabular}
\end{table}

We tested whether explicit chain-of-thought (CoT) prompting improves accuracy on BEAR-Bench. A subset of open-weight models was re-evaluated with a prompt that asks the model to extract information from the image and solve the task step by step before giving a final answer (Table~\ref{tab:cot}; the full prompt is given in Appendix~\ref{app:cot-prompt}). For reasoning-oriented models~---~Qwen3.5-9B, Qwen3.5-4B, and Qwen3-VL-4B-Thinking~---~accuracy is essentially unchanged or slightly lower, consistent with these models already performing intermediate reasoning under the default protocol. By contrast, the instruct-tuned Qwen3-VL-8B-Instruct improves by 13.7 percentage points when steered to externalize multi-step reasoning.

\paragraph{Effect of image resolution.}

\begin{table}[t]
\centering
\caption{Overall accuracy of Qwen3.5-9B on BEAR-Bench at
  different image downsampling factors. A factor of $c$ means
  that both the width and height are divided by $c$.}
\label{tab:image_resolution}
\small
\begin{tabular}{@{}lrrrrr@{}}
\toprule
\textbf{Downsampling factor} & $1$ & $1.5$ & $2$ & $3$ & $4$ \\
\midrule
\textbf{Accuracy (\%)} & 49.6 & 45.0 & 33.0 & 12.6 & 4.7 \\
\bottomrule
\end{tabular}
\end{table}

To measure the sensitivity of document reasoning to image
resolution, we evaluated Qwen3.5-9B after resizing each input
image to $1/c$ of its original width and height, where $c$ is
the downsampling factor in Table~\ref{tab:image_resolution}.
All other inference settings were kept unchanged. Accuracy
decreases from 49.3\% at the original resolution to 45.0\% at
$c=1.5$ and 33.0\% at $c=2$, before falling sharply to 12.6\%
at $c=3$ and 4.7\% at $c=4$. This pronounced degradation
suggests that preserving fine-grained visual detail is
critical for reasoning over text-dense professional
documents.

\subsection{Error Analysis}\label{sec:error_analysis}
To characterize common failure modes on BEAR-Bench, we conducted an exploratory error analysis combining inductive taxonomy construction with manual annotation of model responses.

\subsubsection{Taxonomy construction}
To derive a failure taxonomy grounded in actual model behavior, we first collected an open-coding pilot of 150 incorrect responses produced by Gemini~3.1~Pro on BEAR-Bench, stratified across domains and languages. Four members of our team manually inspected each failure case and wrote detailed, free-form natural-language comments explaining the underlying cause of the error, rather than assigning predefined labels. This yielded a corpus of 150 rich failure descriptions covering a broad range of perceptual and reasoning breakdowns. We then prompted GPT-5.5-Pro, used as an advanced classification assistant, to cluster these free-form comments into thematically coherent groups based on their underlying error mechanism. The resulting clusters were manually reviewed and refined by the authors into five final categories: spatial misgrounding (C1), counting/aggregation (C2), OCR/visual-attribute (C3), chart-value extraction (C4), and semantic/reasoning (C5). Detailed descriptions of each error type are provided in Appendix~\ref{app:error_types}.

\subsubsection{Error statistics}
\begin{figure}[t]
    \centering
    \includegraphics[width=\columnwidth]{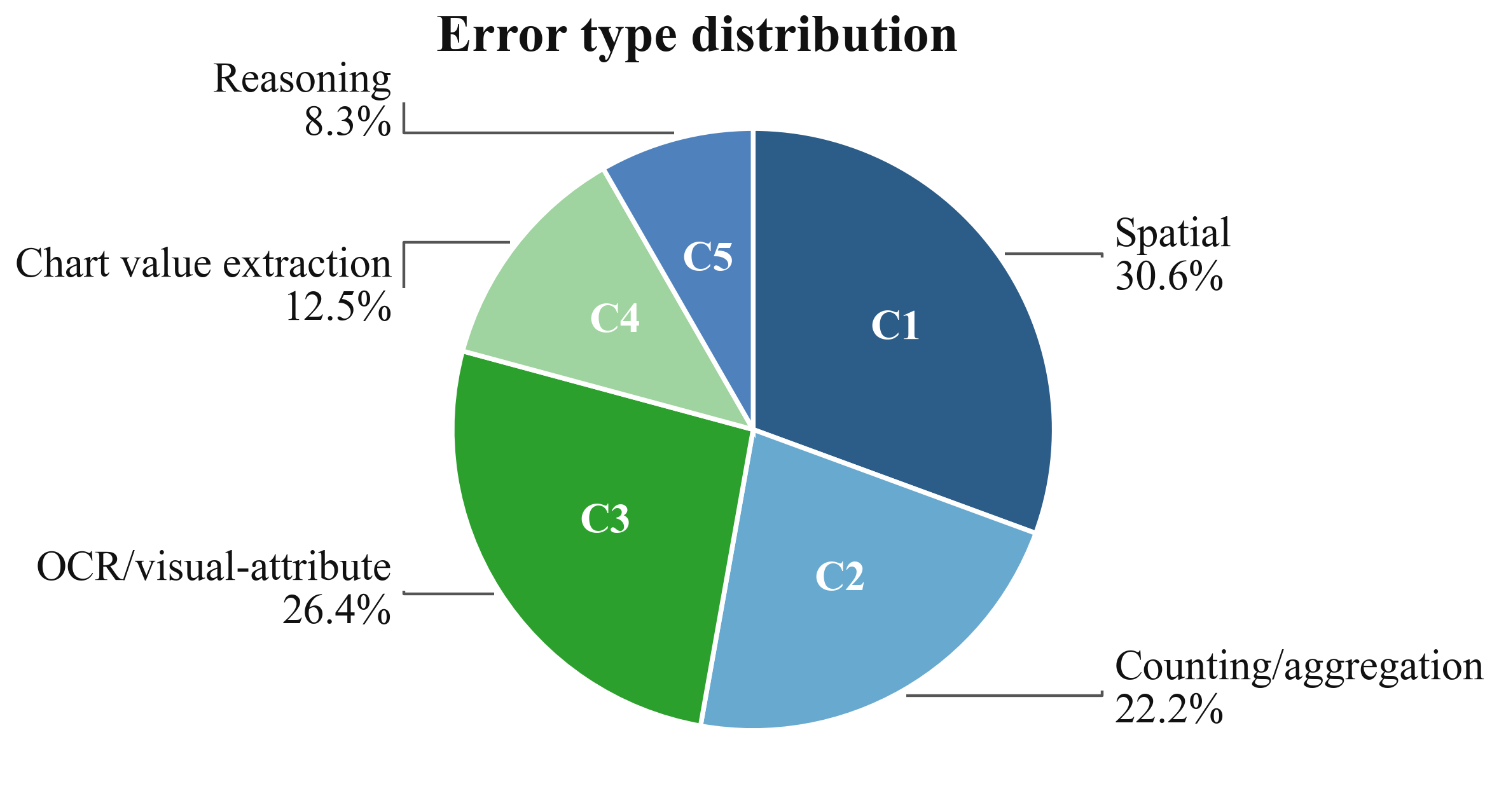}
    \caption{Error-type distribution in a manually annotated subsample of Gemini 3.1 Pro responses ($n=62$). Samples may receive multiple labels. }
    \label{fig:error_types}
\end{figure}
We applied the taxonomy to a subsample of Gemini 3.1 Pro incorrect responses, selected to cover diverse document types and complexity levels. Figure~\ref{fig:error_types} shows the resulting distribution. Visual-perceptual errors dominate: spatial misgrounding (C1) and OCR/visual-attribute errors (C3) together account for the majority of failures. Our analysis suggests that the most common errors on BEAR-Bench are perceptual, involving misread text or visual attributes and mislocalized evidence.


\subsection{Detecting Incorrect Answers}
\label{sec:hallucination}

We next use responses generated on BEAR-Bench to compare
existing hallucination detection methods for text-dense
professional document reasoning. Following the benchmark's
binary answer evaluation, we assess whether each method can
distinguish correct from incorrect responses.

\paragraph{Setup.} We study two open-weight Qwen3.5 models using their native internal signals and eight proprietary models using proxy hidden states extracted from Qwen3-VL-8B~\cite{bai2025qwen3vltechnicalreport}, conditioned on each image, question, and proprietary-model response. We evaluate six uncertainty scores---max/mean token probability, log-likelihood, max/mean entropy, and perplexity---plus ContextualLens~\cite{phukan2025contextuallens}, the supervised hidden-state probe SUQ~\cite{li2024reference}, VIBProbe~\cite{zhang2026vib}, FaithSCAN~\cite{tong2026faithscan}, and an MLLM-as-a-judge baseline (Qwen3-VL-8B)~\cite{gu2025surveyllmasajudge}. We report balanced accuracy (BalAcc), AUROC and AUC-PR metrics.

\paragraph{Results.} The performance of the evaluated methods is shown in Tables~\ref{tab:5fold_gemini}-\ref{tab:5fold_openweight}. No detector wins everywhere; performance depends on the access regime and on response length (Table~\ref{tab:median_response_length_words}). With proxy hidden states for proprietary models, VIB-Probe is the strongest method overall, attaining the best AUROC, AUC-PR, and BalAcc on seven of the eight models (BalAcc $0.71$--$0.76$); FaithScan leads on Gemini~3.1~Flash (BalAcc $0.72$). Among proxy uncertainty scores, max token probability is strongest for the four models answering in two to three words ($0.59$--$0.66$) but is near chance for the four with longer answers ($0.49$--$0.51$), whereas mean-based scores show the reverse pattern ($0.48$--$0.56$ versus $0.65$--$0.66$). With native signals from the two open-weight models, FaithScan achieves the best AUROC and AUC-PR ($0.81$--$0.82$ and $0.77$--$0.82$, respectively), while the MLLM judge obtains the highest BalAcc ($0.80$--$0.81$); both models produce long, detailed responses, making their derivations easier for the judge to verify step by step.

%% file: conclusion.tex
\section{Conclusion}
\label{sec:conclusion}
We introduced BEAR-Bench, a bilingual benchmark of 1{,}000
human-authored questions for context-grounded, multi-step
reasoning over text-dense business and scientific documents.
Benchmark items include dense textual and graphical page content such as
figures, tables, charts, equations, and diagrams. Across 16 proprietary and open-weight MLLMs, the
highest overall accuracy on BEAR-Bench is 75.4\%, and every evaluated model
achieves lower accuracy on Russian items. Our error analysis
suggests that common failures involve spatial
grounding, OCR, and visual-attribute perception.

We also used the resulting model responses to compare existing
hallucination detection methods. Performance varies across
target models: supervised probes perform best for proprietary
outputs, while an MLLM judge achieves the highest balanced
accuracy on the verbose responses from open-weight outputs. The best balanced
accuracy is 0.74 for proprietary outputs and 0.81 for
open-weight models, showing that incorrect responses are not
always identified reliably. BEAR-Bench therefore provides a
common setting for tracking progress in both professional
document reasoning and error detection.

\section*{Limitations}


BEAR-Bench is deliberately narrow in several aspects. All questions are scoped to a single rendered page image; the benchmark therefore does not evaluate multi-page or cross-document reasoning. Coverage is limited to English and Russian enterprise and academic documents drawn from public disclosure and preprint sources, so findings may not transfer to other languages, domains, or private enterprise corpora. With 1{,}000 items and uneven language--domain cell sizes, subset estimates carry more variance than the overall score. Although we validate the LLM-as-a-judge protocol against humans, scoring still depends on an external model, and reasoning-depth labels remain coarse annotator estimates rather than objective difficulty. Finally, our hallucination-detection study compares existing methods under two access regimes and finds that detector quality varies with response length; we do not propose a new detector, and even the best balanced accuracies leave substantial room for improvement.